\documentclass[lettersize,journal]{IEEEtran}
\usepackage{amsmath,amsfonts}
\usepackage{algorithmic}
\usepackage{algorithm}
\usepackage{array}
\usepackage[caption=false,font=normalsize,labelfont=sf,textfont=sf]{subfig}
\usepackage{textcomp}
\usepackage{stfloats}
\usepackage{url}
\usepackage{verbatim}
\usepackage{graphicx}
\usepackage{cite}
\usepackage{amsmath}
\usepackage{booktabs} 
\usepackage{amsmath}
\usepackage{booktabs}
\usepackage{multirow}
\usepackage{graphicx}
\usepackage{pgfplots}
\usepackage{cite}
\begin{document}

\title{Joint Remaining Useful Life Prediction and Capacity Estimation of Lithium-Ion Batteries Using Partial-Charging Data}

\author{Khoa~Tran\textsuperscript{\(\dagger\)},
        Ho-Si-Hung~Nguyen\textsuperscript{*},
        Phone~Wai~Yan~Moe\textsuperscript{\(\dagger\)},
        Hung-Cuong~Trinh,
        and~Thi-Hoang-Giang~Tran%
\thanks{\textsuperscript{\(\dagger\)}Khoa Tran and Phone Wai Yan Moe
contributed equally to this work.}%
\thanks{\textsuperscript{*}Corresponding author:
Ho-Si-Hung Nguyen (e-mail: nhshung@dut.udn.vn).}%
\thanks{Khoa Tran is with Data Science Laboratory,
Faculty of Information Technology, Ton Duc Thang University,
Ho Chi Minh City 70000, Vietnam
(e-mail: trandinhkhoa@tdtu.edu.vn).}%
\thanks{Ho-Si-Hung Nguyen is with Faculty of Electrical
Engineering, The University of Danang - University of Science
and Technology (DUT), 54 Nguyen Luong Bang Street,
Lien Chieu District, Da Nang 550000, Vietnam
(e-mail: nhshung@dut.udn.vn).}%
\thanks{Phone Wai Yan Moe is with AIWARE Limited Company,
17 Huynh Man Dat Street, Hoa Cuong Ward,
Hai Chau District, Da Nang 550000, Vietnam
(e-mail: wyan40653@gmail.com).}%
\thanks{Hung-Cuong Trinh is with Natural Language Processing
and Knowledge Discovery Research Group, Faculty of Information
Technology, Ton Duc Thang University,
Ho Chi Minh City 70000, Vietnam
(e-mail: trinhhungcuong@tdtu.edu.vn).}%
\thanks{Thi-Hoang-Giang Tran is with Faculty of Project
and Industrial Management, The University of Danang - University
of Science and Technology (DUT), 54 Nguyen Luong Bang Street,
Lien Chieu District, Da Nang 550000, Vietnam
(e-mail: tthgiang@dut.udn.vn).}%
}
        % <-this % stops a space



\maketitle

\begin{abstract}
Joint remaining useful life (RUL) prediction and capacity estimation require representations of both gradual degradation and recent battery behavior. This paper presents a cross-expert framework using partial-charging measurements without requiring measured historical full-cycle capacity as an input. The RUL Expert captures long-term degradation from nominal 10-min segments sampled across a 30-cycle history, while the Capacity Expert characterizes recent battery behavior from statistical descriptors of nominal 40-min segments over ten consecutive cycles. Their complementary representations are integrated through feature-wise linear modulation for joint RUL and capacity prediction. A key contribution is a three-stage training strategy that progressively controls frozen and trainable components: supervised representation pretraining, independent expert pretraining, and final fusion training with both experts frozen. This staged optimization preserves expert-specific degradation knowledge while improving the balance between the two prediction tasks, with RUL treated as the primary prognostic objective. On two public battery-aging datasets, the reference configuration achieves mean RUL root-mean-square errors of 143.69 and 161.10 cycles and capacity errors of 12.36 and 7.28~mAh, respectively. On Dataset~I, cross-expert fusion reduces both mean errors relative to either standalone expert. The proposed framework achieves the lowest reported RUL RMSE among the compared methods on both datasets while maintaining competitive capacity-estimation accuracy.
\end{abstract}


\section{Introduction}
\label{sec:introduction}

Lithium-ion batteries (LIBs) are widely used in electric vehicles
(EVs)~\cite{chen2019review} and stationary energy-storage systems because
of their high energy density, long cycle life, and established manufacturing
infrastructure. Transportation electrification and renewable-energy integration
are increasing the demand for reliable battery systems. For example,
\cite{vollert2026assessing} projects that battery demand for e-mobility and
stationary storage in the European Union will reach approximately
$1\,\mathrm{TWh}$ by 2030. Reliable assessment of battery degradation is
therefore important for managing an expanding population of batteries
throughout their service lives.

Battery management systems (BMSs) monitor battery
condition to support operating and maintenance decisions and avoid unnecessary
replacement. Two complementary indicators are state of health
(SOH)~\cite{patrizi2026novel} and remaining
useful life (RUL)~\cite{allal2026machine, zhang2019nonlinear}.
Capacity-based SOH quantifies the available capacity relative to the nominal
capacity, whereas cycle-based RUL describes the remaining cycles before
end of life (EOL):
\begin{align}
    \mathrm{SOH}_c &= \frac{Q_c}{Q_{\mathrm{nom}}}\times100\%,
    \label{eq:soh_definition}\\
    \mathrm{RUL}_c &= c_{\mathrm{EOL}} - c,
    \label{eq:rul_definition}
\end{align}
where $Q_c$ is the available capacity at cycle $c$, $Q_{\mathrm{nom}}$
is the nominal capacity, and $c_{\mathrm{EOL}}$ is the cycle at which
the EOL criterion is reached. Joint capacity estimation and RUL prediction
therefore characterize both the present condition and the anticipated
service life of a battery.

Model-based approaches, including equivalent circuit models
(ECMs)~\cite{chang2022improvement}, relate measured signals
to battery states through parameterized representations. Their performance
depends on model assumptions, parameter identification, and state
initialization, which can become challenging as operating conditions change.
Data-driven methods provide an alternative by learning degradation-related
relationships directly from observations. However, their accuracy and
practical applicability depend on the representativeness of the training data
and the availability of the required measurements during operation.

Recent RUL prediction methods investigate different representations of
battery degradation. Ge et al.~\cite{ge2025deep} introduce a Koopman-inspired
Degradation Model with Embedded Operating Conditions (EKIDM). Charging or
relaxation voltage sequences and operating-condition variables are encoded
into a latent degradation representation, while a condition-dependent Koopman
operator models its evolution. Wu et al.~\cite{wu2025ldnet} use multiple
battery signals within a network employing deformable depthwise convolution.
Learnable sampling offsets adapt feature extraction to the observed signals,
providing a flexible approach to degradation modeling.

Yu et al.~\cite{yu2025multi} extract degradation features from the first
100 complete cycles, primarily using discharge capacity--voltage curves,
$Q(V)$. Features at multiple time scales are selected and processed by a
Transformer-based hybrid network. Although such observations can support
early-life prediction, complete cycling records and reliable discharge-capacity
measurements may be unavailable during routine EV operation. Moreover,
predictions based on early operating conditions may require reassessment
when subsequent usage differs from the training conditions.

For capacity forecasting, Qian et al.~\cite{qian2026transfer} combine
empirical mode decomposition (EMD), a cycle-consistent adversarial network
(CycleGAN), and a condition-focused Transformer. Operating-condition signals
are decomposed into intrinsic mode functions and transformed into
representations used to generate target-condition signals. These signals and
historical capacity sequences are then used for forecasting. Peng et
al.~\cite{peng2026novel} propose the MIC-BOA-TiDE framework, which combines
historical capacity, battery attributes, and selected health indicators.
The maximal information coefficient guides feature selection, a dense
encoder--decoder forecasts capacity, and kernel density estimation provides
prediction intervals from the error distribution. Both approaches illustrate
the usefulness of historical capacity information, but obtaining accurate
full-cycle capacity measurements during routine operation remains challenging.

Partial-charging measurements offer a more accessible source of degradation
information. Ma et al.~\cite{ma2022real} use voltage, accumulated charge,
and their differences within a charging interval from 80\% state of charge
(SOC) to the first occurrence of 3.6~V. An important distinction is whether
accumulated charge is referenced to the beginning of the full charging cycle
or to the start of the observed segment. If the former reference is retained,
values within a partial segment can depend on measurements collected before
that segment. This dependence can complicate deployment when charging begins
at different SOC levels. Segment-local integration addresses this input
requirement by resetting accumulated charge at the beginning of the available
observation, without requiring the preceding charging trajectory.

Song et al.~\cite{song2026uncertainty} further demonstrate the potential of
partial-charging information for SOH estimation. Their framework extracts
the charging time over a specified voltage interval, voltage standard
deviation, and root-mean-square voltage from constant-current charging.
Kernel principal component analysis combines these indicators, and a
Bayesian long short-term memory (LSTM)~\cite{hochreiter1997long} estimates SOH and predictive variance. A collaborative learning
strategy transfers degradation information across operating conditions.
This approach supports the practicality of charging-derived indicators,
while highlighting the value of uncertainty assessment under changing
conditions.

Despite these advances, several limitations remain in existing studies on
battery RUL prediction and capacity/SOH estimation. First, some methods rely
on long observation histories, complete cycling records, or specific charging
protocols, which may limit their applicability when only short
partial-charging measurements are available during routine
operation~\cite{yu2025multi, ma2022real}. Second, many existing approaches
primarily focus on either RUL prediction or capacity/SOH estimation and thus
do not explicitly exploit the complementary temporal information that may
benefit both tasks~\cite{ge2025deep, song2026uncertainty, qian2026transfer}.
Although joint prediction has been investigated~\cite{ma2022real}, effectively
integrating degradation information from different temporal scales for
simultaneous RUL prediction and capacity estimation remains insufficiently
explored. Third, some capacity-forecasting approaches use historical
full-cycle capacity as an input, which may be difficult to obtain reliably
during practical operation~\cite{qian2026transfer, peng2026novel}. Moreover,
accumulated charge obtained directly from a partial charging segment should
be distinguished from historical full-cycle capacity because the two require
different levels of measurement availability~\cite{ma2022real}.

To address these limitations, this study proposes a cross-expert framework for
joint RUL prediction and capacity estimation using only partial-charging
observations. The framework comprises an RUL Expert, a Capacity Expert, and a
feature-wise linear modulation (FiLM) module~\cite{perez2018film}. The RUL
Expert captures long-term degradation behavior by learning cycle-level
representations from short charging segments sampled over an extended cycle
history. In contrast, the Capacity Expert captures more recent degradation
characteristics from statistical descriptors extracted from longer charging
segments over consecutive cycles. Both experts are pretrained using joint RUL
and capacity supervision, while their names reflect their intended temporal
roles rather than exclusive prediction objectives. During cross-expert fusion,
the short-term representation generated by the Capacity Expert produces
feature-wise scaling and shifting parameters that modulate the long-term
representation from the RUL Expert. The fused representation is then passed to
a shared two-output regression head for simultaneous RUL prediction and
capacity estimation. Importantly, neither expert requires historical
full-cycle capacity as an input; both rely on information derived from
partial-charging observations, while RUL and capacity values are used only as
supervised training targets.
The main contributions are summarized as follows:
\begin{enumerate}
    \item \textbf{Complementary partial-charging representations:}
    Nominal 10-min charging segments from ten cycles sampled at stride three
    within a 30-cycle window are paired with nominal 40-min segments from ten
    consecutive cycles. Both views begin at the first observed charging sample
    above 3.1~V. Accumulated charge and elapsed time are referenced to each
    segment's start. Lag-12 differences are computed within the 30-cycle
    window, with unavailable differences set to zero.

    \item \textbf{Cross-expert feature fusion:}
    A pretrained gated recurrent unit (GRU)~\cite{cho2014learning}  encoder, a two-dimensional
    convolutional neural network (2D-CNN)~\cite{gholamrezaii2019human}, and a temporal GRU process the
    long-term view. A 2D-CNN and Transformer process the short-term statistical
    descriptors. FiLM conditions the long-term representation on the
    short-term representation for joint RUL prediction and capacity estimation.

    \item \textbf{Three-stage optimization:}
    Supervised GRU-autoencoder pretraining combines reconstruction with RUL
    and capacity supervision. Independent expert pretraining then retains the
    transferred encoder as a frozen feature extractor. Finally, both experts
    are frozen while the FiLM module and shared regression head are trained.

    \item \textbf{Experimental assessment:}
    Comparisons on two public battery-aging datasets and ablation studies on
    Dataset~I examine input representations, charging-segment duration, cycle
    history, temporal backbones, and fusion strategies. The analysis evaluates
    RUL improvements alongside capacity-estimation trade-offs.
\end{enumerate}

\section{Proposed Method}
\label{sec:proposed_method}

\begin{figure*}[t]
    \centering
    \includegraphics[width=\textwidth,height=0.65\textheight,keepaspectratio]{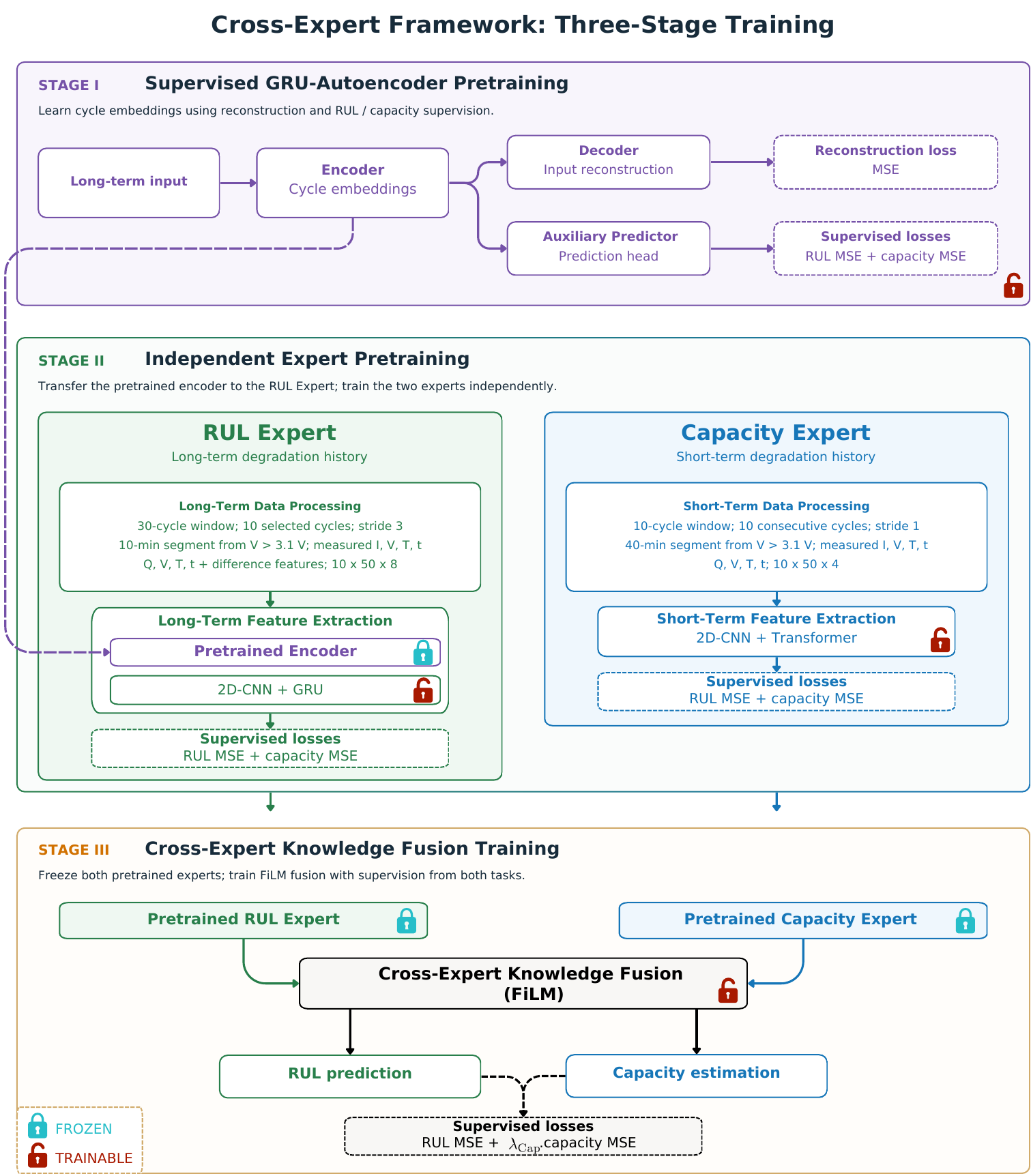}
    \caption{Overview of the proposed cross-expert framework and its three-stage
    training procedure. Stage I learns cycle-level embeddings through supervised
    GRU-autoencoder pretraining. Stage II independently trains the RUL and
    Capacity Experts using long- and short-term degradation histories,
    respectively. Stage III freezes both experts and optimizes the FiLM-based
    fusion module and shared regression head for joint RUL prediction and
    capacity estimation. Tensor dimensions in the diagram omit the batch axis.}
    \label{fig:Overall_flow}
\end{figure*}

As illustrated in Fig.~\ref{fig:Overall_flow}, the proposed framework
comprises an RUL Expert, a Capacity Expert, and a feature-wise linear
modulation (FiLM) module for cross-expert knowledge fusion. The two
experts capture complementary temporal contexts. The RUL Expert
processes nominal 10-min partial charging segments sampled from a
30-cycle window to characterize gradual degradation. Lag-12
inter-cycle differences are computed within this window. The Capacity Expert processes nominal 40-min charging
segments from the 10 most recent consecutive cycles to characterize
short-term battery behavior. The framework is optimized in three
stages: 1) supervised GRU-autoencoder pretraining,
2) independent multi-task pretraining of both experts, and
3) FiLM-based fusion training with both experts frozen.

\subsection{RUL Expert}
\label{subsec:rul_expert}

The RUL Expert characterizes degradation over an extended cycle history. Its
pipeline consists of long-term data preparation, a pretrained GRU encoder for
cycle-level representation learning, a 2D-CNN for local feature extraction, and
a temporal GRU for modeling degradation across cycles.

\subsubsection{Long-Term Data Processing}

For cycle $c$, the measured current, voltage, and temperature sequences are
denoted by $\mathbf{I}_c$, $\mathbf{V}_c$, and $\mathbf{T}_c$, with corresponding
timestamps $\mathbf{t}_c$. Bold symbols denote sequences, whereas $I_{c,j}$,
$V_{c,j}$, and $T_{c,j}$ denote individual samples. A nominal 10-min charging
segment is extracted beginning at the first sample satisfying $V>3.1\,\mathrm{V}$. Using partial charging segments
supports practical battery-management applications in which complete
charge--discharge records may be unavailable.

Let $M_c$ be the number of samples in the extracted segment. A second-order
Savitzky--Golay filter~\cite{savitzky1964smoothing} $\mathcal{G}(\cdot)$ is independently applied to current,
voltage, and temperature:
\begin{equation}
\begin{aligned}
    \bar{\mathbf{I}}_c &= \mathcal{G}(\mathbf{I}_c), &
    \bar{\mathbf{V}}_c &= \mathcal{G}(\mathbf{V}_c), &
    \bar{\mathbf{T}}_c &= \mathcal{G}(\mathbf{T}_c).
\end{aligned}
\label{eq:sg_filter}
\end{equation}
The odd filter window is selected as approximately 25\% of $M_c$ and is clipped
to satisfy the segment-length and polynomial-order constraints.

Elapsed time is measured relative to the beginning of the segment:
\begin{equation}
    \tau_{c,j}=t_{c,j}-t_{c,0},
    \qquad j=0,\ldots,M_c-1.
\label{eq:elapsed_time}
\end{equation}
The charge accumulated within the observed segment is obtained by numerical
integration of the filtered current:
\begin{equation}
\begin{aligned}
    Q^{L}_{c,0} &= 0,\\
    Q^{L}_{c,j} &= \sum_{k=1}^{j}\bar{I}_{c,k}
    \left(\tau_{c,k}-\tau_{c,k-1}\right),
    \quad j=1,\ldots,M_c-1.
\end{aligned}
\label{eq:accumulated_charge}
\end{equation}
Current and time are expressed in compatible units, such that amperes and hours
yield ampere-hours. Here, $Q^{L}_{c,j}$ is the charge accumulated only within
the observed partial segment and is distinct from the full-cycle capacity used
as a prediction target.

The four long-term channels are assembled as
\begin{equation}
    \mathbf{Z}^{L}_c =
    \begin{bmatrix}
        \mathbf{Q}^{L}_c & \bar{\mathbf{V}}_c &
        \bar{\mathbf{T}}_c & \boldsymbol{\tau}_c
    \end{bmatrix}
    \in\mathbb{R}^{M_c\times4}.
\label{eq:long_channels}
\end{equation}
Each channel is linearly interpolated over sample positions to a
fixed length of 50:
\begin{equation}
    \mathbf{X}^{L}_c=\mathcal{I}_{50}(\mathbf{Z}^{L}_c)
    \in\mathbb{R}^{50\times4},
\label{eq:long_interpolation}
\end{equation}
where $\mathcal{I}_{50}(\cdot)$ denotes channel-wise interpolation.

To explicitly represent degradation changes, a 12-cycle lag difference is
computed within each 30-cycle observation window. Let
$\mathbf{X}^{L}_{w,p}$ denote the representation at chronological position
$p\in\{0,\ldots,29\}$ in window $w$.
Then
\begin{equation}
    \Delta\mathbf{X}^{L}_{w,p}=
    \begin{cases}
        \mathbf{0}, & p<12,\\
        \mathbf{X}^{L}_{w,p}-\mathbf{X}^{L}_{w,p-12}, & p\geq12.
    \end{cases}
\label{eq:long_difference}
\end{equation}
The first 12 difference tensors are zero because their reference cycles
precede the window. No observations outside the 30-cycle window are used.
The base and difference channels are concatenated as
\begin{equation}
    \widetilde{\mathbf{X}}^{L}_{w,p}
    =\operatorname{Concat}_{\mathrm{ch}}
    \left(\mathbf{X}^{L}_{w,p},\Delta\mathbf{X}^{L}_{w,p}\right)
    \in\mathbb{R}^{50\times8}.
\label{eq:long_augmented}
\end{equation}

Min--max normalization $\mathcal{M}_{L}$ is applied independently to each
resampled-position--channel pair. Its parameters are estimated exclusively from the
training set and subsequently reused for validation and testing:
\begin{equation}
    \widehat{\mathbf{X}}^{L}_{w,p}
    =\mathcal{M}_{L}\!\left(\widetilde{\mathbf{X}}^{L}_{w,p}\right)
    \in\mathbb{R}^{50\times8}.
\label{eq:long_scaling}
\end{equation}
Ten cycles are selected at stride three from the 30-cycle window, i.e.,
$p_k=3k$ for $k=0,\ldots,9$. Thus, positions 28 and 29 are not selected
for this branch. The long-term model input is therefore
\begin{equation}
\begin{aligned}
    \mathbf{H}^{L}_w
    &=\operatorname{Stack}_{\mathrm{cycle}}
    \left(
        \widehat{\mathbf{X}}^{L}_{w,0},
        \widehat{\mathbf{X}}^{L}_{w,3},\ldots,
        \widehat{\mathbf{X}}^{L}_{w,27}
    \right)\\
    &\in\mathbb{R}^{10\times50\times8}.
\end{aligned}
\label{eq:long_input}
\end{equation}

\subsubsection{GRU-Autoencoder}
\label{subsubsec:gru_autoencoder}

The GRU-autoencoder learns compact cycle-level representations
from partial charging measurements. Its encoder maps each
$50\times8$ input sequence to a 64-dimensional embedding.
The decoder projects this embedding to 128 dimensions and
repeats it across 50 sequence positions. A two-layer GRU
followed by a linear output layer reconstructs the eight
input channels at each position. Reconstruction is
non-autoregressive: previously reconstructed samples are
not fed back into the decoder.

The embeddings from 10 sampled cycles are also stacked
and processed by an auxiliary GRU predictor for joint
RUL prediction and capacity estimation.
Table~\ref{tab:autoencoder_arch} summarizes the architecture.

During Stage I, the encoder, decoder, and auxiliary predictor
are optimized jointly using reconstruction and prediction
losses. This supervision encourages the embeddings to retain
input information relevant to battery degradation. After
pretraining, only the encoder is transferred to the RUL
Expert and frozen; the decoder and auxiliary predictor
are discarded.

\begin{table}[t]
\centering
\caption{Supervised GRU-Autoencoder Pretraining Architecture}
\label{tab:autoencoder_arch}
\footnotesize
\setlength{\tabcolsep}{3pt}
\renewcommand{\arraystretch}{1.10}
\resizebox{\columnwidth}{!}{%
\begin{tabular}{@{}lll@{}}
\hline
\textbf{Component} &
\textbf{Operation} &
\textbf{Output shape} \\
\hline
Input
& Long-term window
& $B\times10\times50\times8$ \\
& Merge batch/cycle axes
& $10B\times50\times8$ \\
\hline
Encoder
& GRU, 2 layers
& $10B\times50\times128$ \\
& Final top-layer state
& $10B\times128$ \\
& Linear, $128\rightarrow64$
& $10B\times64$ \\
\hline
Decoder
& Linear, $64\rightarrow128$
& $10B\times128$ \\
& Repeat over 50 steps
& $10B\times50\times128$ \\
& GRU, 2 layers
& $10B\times50\times128$ \\
& Linear, $128\rightarrow8$
& $10B\times50\times8$ \\
\hline
Auxiliary
& Restore cycle axis
& $B\times10\times64$ \\
predictor
& GRU, 2 layers
& $B\times10\times128$ \\
& Final top-layer state
& $B\times128$ \\
& Two-output regression head
& $B\times2$ \\
\hline
\end{tabular}%
}

\smallskip
\parbox{\columnwidth}{\footnotesize
$B$ denotes the mini-batch size. All three GRUs use
128 hidden units per layer and inter-layer dropout
of 0.1. The decoder and auxiliary predictor form
separate branches receiving the encoder embeddings.
The decoder reconstructs each cycle independently,
whereas the predictor models the sequence of 10 cycle
embeddings to estimate RUL and capacity.}
\end{table}

\subsubsection{Long-Term Degradation Feature Extraction}
\label{subsubsec:long_feature_extraction}

The pretrained encoder independently processes the 10 cycle
sequences in $\mathbf{H}^{L}_{w}$. Their embeddings are stacked
into a $10\times64$ degradation map, which is processed by
two 2D convolutional layers. Mean pooling over the
embedding-position axis produces a sequence of 10
cycle-level feature vectors, each with 128 dimensions.

A two-layer temporal GRU models dependencies across the
sampled cycles. Its final top-layer hidden state,
$\mathbf{h}^{L}_{w}\in\mathbb{R}^{128}$, is supplied to the
two-output regression head during Stage II and to the
FiLM fusion module during Stage III.
Table~\ref{tab:long_arch} summarizes the architecture.

\begin{table}[t]
\centering
\caption{RUL Expert Architecture}
\label{tab:long_arch}
\footnotesize
\setlength{\tabcolsep}{3pt}
\renewcommand{\arraystretch}{1.10}
\resizebox{\columnwidth}{!}{%
\begin{tabular}{@{}lll@{}}
\hline
\textbf{Component} &
\textbf{Operation} &
\textbf{Output shape} \\
\hline
Input
& Long-term window
& $B\times10\times50\times8$ \\
& Merge batch/cycle axes
& $10B\times50\times8$ \\
\hline
Pretrained
& GRU encoder, final state
& $10B\times128$ \\
Encoder
& Linear, $128\rightarrow64$
& $10B\times64$ \\
& Restore cycle axis
& $B\times10\times64$ \\
\hline
Long-Term
& Add channel axis
& $B\times1\times10\times64$ \\
Degradation
& Conv2D, $1\rightarrow32$
& $B\times32\times10\times64$ \\
Feature Extraction
& Conv2D, $32\rightarrow128$
& $B\times128\times10\times64$ \\
& Mean over embedding axis
& $B\times128\times10$ \\
& Transpose
& $B\times10\times128$ \\

& Temporal GRU, final state
& $B\times128$ \\
\hline
Predictor
& Two-output regression head
& $B\times2$ \\
\hline
\end{tabular}%
}

\smallskip
\parbox{\columnwidth}{\footnotesize
Both GRUs have two layers with 128 hidden
units per layer and inter-layer dropout of 0.1.
Both convolutions use $3\times3$ kernels, unit stride,
and padding of one. Each convolution is followed by
ReLU; dropout of 0.1 follows the first ReLU.
The transferred encoder remains frozen during Stage II,
while the expert CNN, temporal GRU, and regression head
are trained. Stage III uses the 128-dimensional expert
representation before the regression head.}
\end{table}

\subsection{Capacity Expert}
\label{subsec:capacity_expert}

The Capacity Expert characterizes recent battery behavior
using longer partial charging segments from 10 consecutive
cycles. Channel-wise statistical descriptors summarize
each segment, and a 2D-CNN followed by a Transformer encoder
models the resulting cycle sequence.

\subsubsection{Short-Term Data Processing}

For each cycle, a nominal 40-min charging segment is extracted beginning at the
first sample satisfying $V>3.1\,\mathrm{V}$. Savitzky--Golay filtering is not applied in this branch. Elapsed time and
accumulated charge are calculated using
Eqs.~\eqref{eq:elapsed_time} and \eqref{eq:accumulated_charge}, respectively,
with the original current replacing the filtered current and the summation
taken over the 40-min segment. Its locally accumulated charge is denoted by
$Q^{S}_{c,j}$ to distinguish it from $Q^{L}_{c,j}$. Raw-signal and elapsed-time
symbols in this subsection refer to the short-term segment. The resulting
channels are
\begin{equation}
    \mathbf{Z}^{S}_c =
    \begin{bmatrix}
        \mathbf{Q}^{S}_c & \mathbf{V}_c &
        \mathbf{T}_c & \boldsymbol{\tau}_c
    \end{bmatrix}
    \in\mathbb{R}^{M_c^{S}\times4},
\label{eq:short_channels}
\end{equation}
where $M_c^{S}$ denotes the number of samples in the extracted short-term
segment. Let $p\in\{0,\ldots,9\}$ index its chronological cycle position
within the short-term window associated with sample $w$.
Channel-wise interpolation and training-set min--max normalization
produce
\begin{equation}
    \widehat{\mathbf{X}}^{S}_{w,p}
    =\mathcal{M}_{S}\!\left(
        \mathcal{I}_{50}(\mathbf{Z}^{S}_{w,p})
    \right)
    \in\mathbb{R}^{50\times4}.
\label{eq:short_scaled}
\end{equation}
The normalization parameters are fitted only on the training set and are reused
without modification for validation and testing.

For each channel $q\in\{1,\ldots,4\}$, seven descriptors are computed over the
50 normalized, resampled positions:
\begin{equation}
    \mathbf{d}^{S}_{w,p,q}
    =\left[
        \mu,\,\sigma,\,\min,\,\max,\,
        \operatorname{median},\,\sigma^{2},\,
        \operatorname{skew}
    \right]^{\top}
    \in\mathbb{R}^{7}.
\label{eq:short_descriptors}
\end{equation}
Here, $\mu$, $\sigma$, and $\sigma^2$ denote the mean, standard deviation,
and variance, respectively; the remaining entries denote the minimum,
maximum, median, and skewness of the same channel sequence.
Concatenating the descriptors of all four channels yields the cycle-level
statistical vector
\begin{equation}
    \mathbf{s}^{S}_{w,p}
    =\operatorname{Concat}
    \left(
        \mathbf{d}^{S}_{w,p,1},\ldots,
        \mathbf{d}^{S}_{w,p,4}
    \right)
    \in\mathbb{R}^{28}.
\label{eq:short_vector}
\end{equation}
The vectors from 10 consecutive cycles, sampled at stride one, are stacked to
form the Capacity Expert input:
\begin{equation}
    \mathbf{A}^{S}_w
    =\operatorname{Stack}_{\mathrm{cycle}}
    \left(
        \mathbf{s}^{S}_{w,0},\ldots,
        \mathbf{s}^{S}_{w,9}
    \right)
    \in\mathbb{R}^{10\times28}.
\label{eq:short_input}
\end{equation}

\subsubsection{Short-Term Degradation Feature Extraction}

The Capacity Expert processes the statistical descriptor matrix
$\mathbf{A}^{S}_w\in\mathbb{R}^{10\times28}$ using a 2D-CNN
followed by a Transformer encoder~\cite{vaswani2017attention}. The CNN preserves the cycle
and statistical-feature axes while expanding the channel
dimension to 128. Averaging over the statistical-feature axis
and transposing the result produces a sequence of 10 cycle
tokens, each with 128 features.

A learned linear projection maps these tokens to the
Transformer embedding space, and fixed sinusoidal positional
encoding represents their order within the cycle window.
The resulting sequence is processed by two Transformer
encoder layers. Mean pooling over the cycle axis yields
the short-term representation
$\mathbf{h}^{S}_w\in\mathbb{R}^{128}$, which is passed to
a two-output regression head for joint RUL prediction
and capacity estimation during Stage II. During Stage III,
this representation is extracted before the regression
head and supplied to the fusion module.
Table~\ref{tab:short_arch} summarizes the architecture.

\begin{table}[t]
\centering
\caption{Capacity Expert Architecture}
\label{tab:short_arch}
\footnotesize
\setlength{\tabcolsep}{3pt}
\renewcommand{\arraystretch}{1.10}
\resizebox{\columnwidth}{!}{%
\begin{tabular}{@{}lll@{}}
\hline
\textbf{Component} &
\textbf{Operation} &
\textbf{Output shape} \\
\hline
Input
& Statistical descriptors
& $B\times10\times28$ \\
\hline
Short-Term
& Add channel axis
& $B\times1\times10\times28$ \\
Degradation
& Conv2D, $1\rightarrow32$
& $B\times32\times10\times28$ \\
Feature Extraction
& Conv2D, $32\rightarrow128$
& $B\times128\times10\times28$ \\
& Mean over feature axis
& $B\times128\times10$ \\
& Transpose to cycle sequence
& $B\times10\times128$ \\
& Linear, $128\rightarrow128$
& $B\times10\times128$ \\
& Add sinusoidal encoding
& $B\times10\times128$ \\
& Transformer encoder, 2 layers
& $B\times10\times128$ \\
& Mean over cycle tokens
& $B\times128$ \\
\hline
Predictor
& Two-output regression head
& $B\times2$ \\
\hline
\end{tabular}%
}

\smallskip
\parbox{\columnwidth}{\footnotesize
$B$ denotes the mini-batch size. Both convolutions use
$3\times3$ kernels, unit stride, and padding of one,
followed by ReLU; dropout of 0.1 follows the first ReLU.
The Transformer has an embedding dimension of 128,
four attention heads, a feedforward dimension of 512,
and dropout of 0.1. Fixed sinusoidal positional encoding
supports up to 10 cycle tokens. The pooled representation
is extracted before the prediction head for Stage III.}
\end{table}

\subsection{Cross-Expert Knowledge Fusion}
\label{subsec:fusion}

The fusion module integrates the long-term representation
$\mathbf{h}^{L}_w\in\mathbb{R}^{128}$ and the short-term
representation $\mathbf{h}^{S}_w\in\mathbb{R}^{128}$ through
FiLM. The short-term representation generates feature-wise
scaling and shifting vectors that modulate the long-term
representation:
\begin{equation}
    \widetilde{\mathbf{h}}_w
    = \boldsymbol{\gamma}(\mathbf{h}^{S}_w)
      \odot \mathbf{h}^{L}_w
    + \boldsymbol{\beta}(\mathbf{h}^{S}_w),
    \label{eq:film}
\end{equation}
where $\odot$ denotes element-wise multiplication, and
$\boldsymbol{\gamma},\boldsymbol{\beta}:
\mathbb{R}^{128}\rightarrow\mathbb{R}^{128}$ are learned
mappings. This modulation allows recent battery behavior
to condition the long-term degradation representation.

The fused representation $\widetilde{\mathbf{h}}_w$ is
passed to a shared regression head comprising a linear
layer, ReLU activation, dropout, and a final linear layer
with two outputs for RUL prediction and capacity estimation:
\begin{equation}
    \widehat{\mathbf{y}}_w
    = \mathbf{W}_2
      \operatorname{Dropout}\!\left(
      \operatorname{ReLU}\!\left(
      \mathbf{W}_1\widetilde{\mathbf{h}}_w+\mathbf{b}_1
      \right)\right)
      +\mathbf{b}_2
    \in\mathbb{R}^{2},
    \label{eq:fusion_head}
\end{equation}
where $\mathbf{W}_1\in\mathbb{R}^{d_h\times128}$ and
$\mathbf{W}_2\in\mathbb{R}^{2\times d_h}$ are learned
weight matrices, $\mathbf{b}_1$ and $\mathbf{b}_2$ are
bias vectors with dimensions $d_h$ and 2, respectively, and $d_h$ denotes
the head's hidden dimension. The two entries of
$\widehat{\mathbf{y}}_w$ are the scaled RUL and capacity estimates.
Physical units are restored before evaluation as described in
Section~\ref{subsec:evaluation_metrics}.

During Stage III, both pretrained experts remain frozen.
The frozen experts are used in evaluation mode to disable dropout.
Only the FiLM module and shared regression head are
optimized, preserving the expert representations while
learning their interaction for joint prediction.

\subsection{Three-Stage Training Strategy}
\label{subsec:training_strategy}

Training comprises three sequential stages: supervised
GRU-autoencoder pretraining, independent expert pretraining,
and cross-expert fusion training.

The RUL and capacity targets are scaled as
\begin{equation}
    \widetilde{y}_{\mathrm{RUL}}
    = \frac{y_{\mathrm{RUL}}}{s_{\mathrm{RUL}}},
    \qquad
    \widetilde{y}_{\mathrm{Cap}}
    = \frac{y_{\mathrm{Cap}}}{s_Q},
\end{equation}
where $s_{\mathrm{RUL}}=3000$ cycles,
$s_Q=1.1\,\mathrm{Ah}$ for Dataset~I, and
$s_Q=1300\,\mathrm{mAh}$ for Dataset~II.
The capacity scaling factors match the native units
of the corresponding datasets.

Let $\widetilde{\mathbf{y}}_{\mathrm{RUL}}$ and
$\widetilde{\mathbf{y}}_{\mathrm{Cap}}$ denote the scaled
target vectors within a mini-batch, and let
$\widehat{\widetilde{\mathbf{y}}}_{\mathrm{RUL}}$ and
$\widehat{\widetilde{\mathbf{y}}}_{\mathrm{Cap}}$ denote
the corresponding model outputs in the scaled domain.
The task and reconstruction losses are defined as
\begin{equation}
\begin{aligned}
    \mathcal{L}_{\mathrm{RUL}}
    &= \operatorname{MSE}
    \left(
        \widehat{\widetilde{\mathbf{y}}}_{\mathrm{RUL}},
        \widetilde{\mathbf{y}}_{\mathrm{RUL}}
    \right), \\
    \mathcal{L}_{\mathrm{Cap}}
    &= \operatorname{MSE}
    \left(
        \widehat{\widetilde{\mathbf{y}}}_{\mathrm{Cap}},
        \widetilde{\mathbf{y}}_{\mathrm{Cap}}
    \right), \\
    \mathcal{L}_{\mathrm{rec}}
    &= \operatorname{MSE}
    \left(
        \widehat{\mathbf{H}}^{L},
        \mathbf{H}^{L}
    \right).
\end{aligned}
\label{eq:base_losses}
\end{equation}
Here, $\mathbf{H}^{L}$ denotes the mini-batch of
preprocessed long-term inputs, and
$\widehat{\mathbf{H}}^{L}$ denotes its reconstruction
after restoring the batch and cycle axes.
Each MSE is averaged over all elements of the
corresponding tensors.

\paragraph{Stage I: Supervised GRU-autoencoder pretraining}
The GRU-based autoencoder and its auxiliary GRU
prediction backbone are jointly pretrained for
100 epochs using
\begin{equation}
    \mathcal{L}_{\mathrm{I}}
    =
    \mathcal{L}_{\mathrm{RUL}}
    +
    \mathcal{L}_{\mathrm{Cap}}
    +
    \mathcal{L}_{\mathrm{rec}}.
\end{equation}
The reconstruction term encourages preservation of
the observed signal structure, while the task losses
encourage representations informative for both RUL
prediction and capacity estimation.

\paragraph{Stage II: Independent expert pretraining}
The pretrained GRU encoder is transferred to the
RUL Expert and kept frozen. The remaining RUL Expert
components and the Capacity Expert are trained
independently for 5 epochs each. Both experts use
joint RUL and capacity supervision:
\begin{equation}
    \mathcal{L}_{\mathrm{II}}
    =
    \mathcal{L}_{\mathrm{RUL}}
    +
    \mathcal{L}_{\mathrm{Cap}}.
\end{equation}
The expert names reflect their respective temporal
representations rather than exclusive prediction
objectives. No reconstruction loss is used in this stage.

\paragraph{Stage III: Cross-expert fusion training}
Both pretrained experts are frozen, and the FiLM
fusion module and shared two-output regression head
are trained for 5 epochs using
\begin{equation}
    \mathcal{L}_{\mathrm{III}}
    =
    \mathcal{L}_{\mathrm{RUL}}
    +
    \lambda_{\mathrm{Cap}}\mathcal{L}_{\mathrm{Cap}}.
\end{equation}
Only the fusion module and regression head are updated,
preserving the representations learned by the experts.

All three stages use the Adam optimizer~\cite{kingma2014adam}
with a learning rate of $10^{-4}$ and a mini-batch size of 128.
Hyperparameters are selected through grid
search~\cite{liashchynskyi2019grid}, with hidden dimensions
in $\{64, 128, 256, 384\}$, dropout rates in
$\{0, 0.1, 0.2, 0.3\}$, and numbers of layers in
$\{1, 2, 3, 4\}$.
Predictions are converted back to physical units before
computing the evaluation metrics defined in
Section~\ref{subsec:evaluation_metrics}. Capacity values are converted to mAh
for consistent reporting across datasets.

\section{Experimental Setup}
\label{sec:experimental_setup}

\subsection{Datasets}
Two public battery-aging datasets are considered. Dataset~I, released by \cite{severson2019data}, contains 124 commercial
A123 APR18650M1A lithium iron phosphate (LFP)/graphite cells, with a nominal
capacity of 1.1~Ah and a nominal voltage of 3.3~V. The cells were cycled at
$30\,^{\circ}\mathrm{C}$ under different fast-charging protocols, yielding cycle
lives of approximately 150--2300 cycles. The dataset includes electrical and
surface-temperature measurements and provides a benchmark for degradation
prediction under charging-protocol variability.

Dataset~II, released by Ma et al.~\cite{ma2022real}, comprises 77 commercial
LFP/graphite cells with a nominal capacity of 1.1~Ah. The cells were tested at
$30\,^{\circ}\mathrm{C}$ using an identical charging procedure and 77 different
multistage discharge protocols. The charging procedure includes 5C charging
to 80\% state of charge, followed by 1C charging to 3.6~V and a
constant-voltage stage. The dataset contains more than 140\,000 cycles, with
cell lifetimes of approximately 1100--2700 cycles. It complements Dataset~I
by emphasizing discharge-protocol variability. Both source studies use 80\% of nominal capacity as the end-of-life criterion.

\subsection{Data Partitioning}
\label{subsec:evaluation_protocol}

Both datasets are partitioned by battery identity, ensuring that
all windows from a given cell belong to the same partition.

For Dataset~I, we adopt the cell grouping reported in the
supplementary information of \cite{severson2019data}, using
41 cells for training, 43 for validation, and 40 for testing.
The source study's primary test set is used here for validation,
while its secondary test set serves as the test set. Training and validation
cells come from the 2017-05-12 and 2017-06-30 batches; the 40 test cells
come from the later 2018-04-12 batch. Thus, Dataset~I evaluates unseen
cells under a manufacturing-batch and charging-protocol shift.

For Dataset~II, we follow the predefined division of
55 development cells and 22 test cells in \cite{ma2022real}.
The 55 development cells are further divided into training
and validation subsets using an approximately 90:10 ratio
at the cell level. The remaining 22 cells are reserved
exclusively for final testing.

All learned preprocessing parameters are fitted using only
the training partition and remain fixed during validation
and testing. Rolling windows contain only measurements
available at the prediction time. Model configurations,
loss weights, and checkpoints are selected using the
validation partition; the test partitions are reserved
for final performance evaluation.

\subsection{Implementation Details}
\label{subsec:implementation}

The neural networks are implemented in PyTorch~2.5.1.
The training procedure and hyperparameter search are
described in Section~\ref{subsec:training_strategy},
and the selected architectures are reported in the
corresponding model-configuration tables.

Data preprocessing and numerical computations use
NumPy~2.2.6, SciPy~1.15.3, pandas~2.3.3, and
scikit-learn~1.7.2. XGBoost~3.2.0 is used for the
tree-based regression experiments. All experiments are
conducted on a workstation equipped with an NVIDIA
GeForce RTX~5060 GPU, an AMD Ryzen~7 CPU, and 32~GB
of system RAM.

\subsection{Evaluation Metrics}
\label{subsec:evaluation_metrics}

Performance is evaluated separately for RUL prediction
and capacity estimation using root-mean-square error
(RMSE), the coefficient of determination ($R^2$), and
mean absolute percentage error (MAPE).
Predictions and targets are first restored to physical
units, with RUL expressed in cycles and capacity in mAh
for both datasets.

For a test cell $c$ containing $N_c$ evaluated windows,
the metrics are defined as
\begin{align}
    \mathrm{RMSE}_c
    &= \sqrt{\frac{1}{N_c}
       \sum_{i=1}^{N_c}(y_{c,i}-\hat{y}_{c,i})^2}, \\
    R_c^2
    &= 1-
       \frac{\sum_{i=1}^{N_c}(y_{c,i}-\hat{y}_{c,i})^2}
       {\sum_{i=1}^{N_c}(y_{c,i}-\bar{y}_c)^2}, \\
    \mathrm{MAPE}_c
    &= \frac{100}{N_c}
       \sum_{i=1}^{N_c}
       \left|\frac{y_{c,i}-\hat{y}_{c,i}}{y_{c,i}}\right|,
\end{align}
where $y_{c,i}$ and $\hat{y}_{c,i}$ denote the true
and predicted values for window $i$ of test cell $c$,
respectively, and
$\bar{y}_c=N_c^{-1}\sum_{i=1}^{N_c}y_{c,i}$.
The task subscript is omitted for clarity.
RMSE is reported in cycles for RUL and in mAh for
capacity, whereas MAPE is reported as a percentage.
Lower RMSE and MAPE and higher $R^2$ indicate better
predictive performance.
MAPE requires nonzero targets, and $R^2$ requires
a nonconstant target sequence.

% Each metric is computed separately for each test cell
% and then averaged with equal weight across test cells,
% rather than pooling all evaluation windows.

Predictions are averaged over five runs using
fixed data partitions. This prediction-level averaging forms a five-model
ensemble. Evaluation metrics are then computed separately for
each test cell and summarized as the mean and standard deviation across
test cells, $\bar{m}\pm s_m$. The standard deviation therefore quantifies
between-cell variability in predictive performance, rather than variability
across runs.

\subsection{Baseline Methods}

The evaluation comprises three groups of controlled
comparisons and additional results reported in prior studies.

First, GRU, LSTM, Transformer,
Ge2025~\cite{ge2025deep}, and
Qian2026~\cite{qian2026transfer} are rerun in this study.
The baselines
process nominal 40-min charging segments beginning at
the first charging sample above 3.1~V, without
autoencoder embeddings or handcrafted statistics.

Second, all signal-processing ablation studies use
XGBoost~\cite{chen2016xgboost} as the downstream predictor,
selected to provide stable training and reliable prediction.
A common predictor and consistent training protocol
support a fair comparison of statistical descriptors,
raw resampled curves, and learned embeddings.

Third, fusion ablations compare alternative feature-
or prediction-level fusion operators using the same
two frozen pretrained experts.

Results for MSFEH~\cite{yu2025multi}, the structural
pruning method~\cite{ge2024structural}, and
Ma2022~\cite{ma2022real} are taken directly from their
original publications, which report mean metrics without
standard deviations across test cells.
Accordingly, Tables~\ref{tab:sota_comparison_datasets}
and~\ref{tab:main_results_verified} report only mean
values for all methods to maintain a consistent
presentation.

\section{Results and Discussion}
\label{sec:results}

\begin{figure*}[t]
\centering
\includegraphics[width=0.48\textwidth]{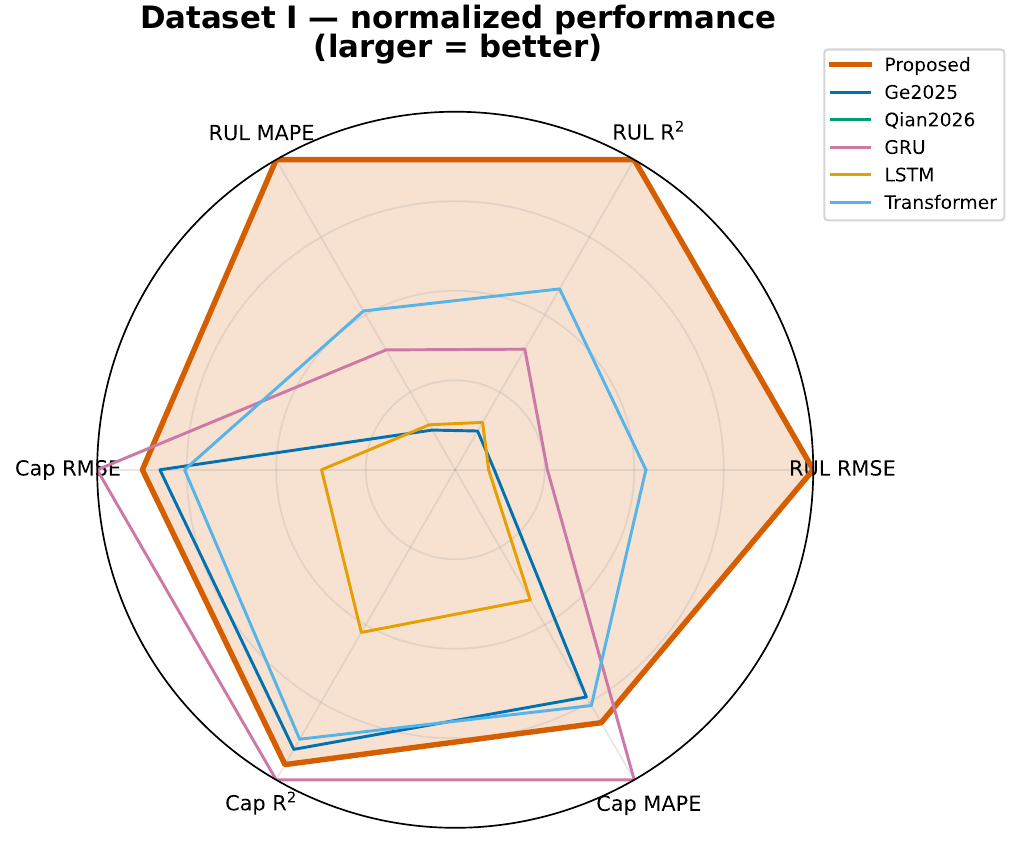}\hfill
\includegraphics[width=0.48\textwidth]{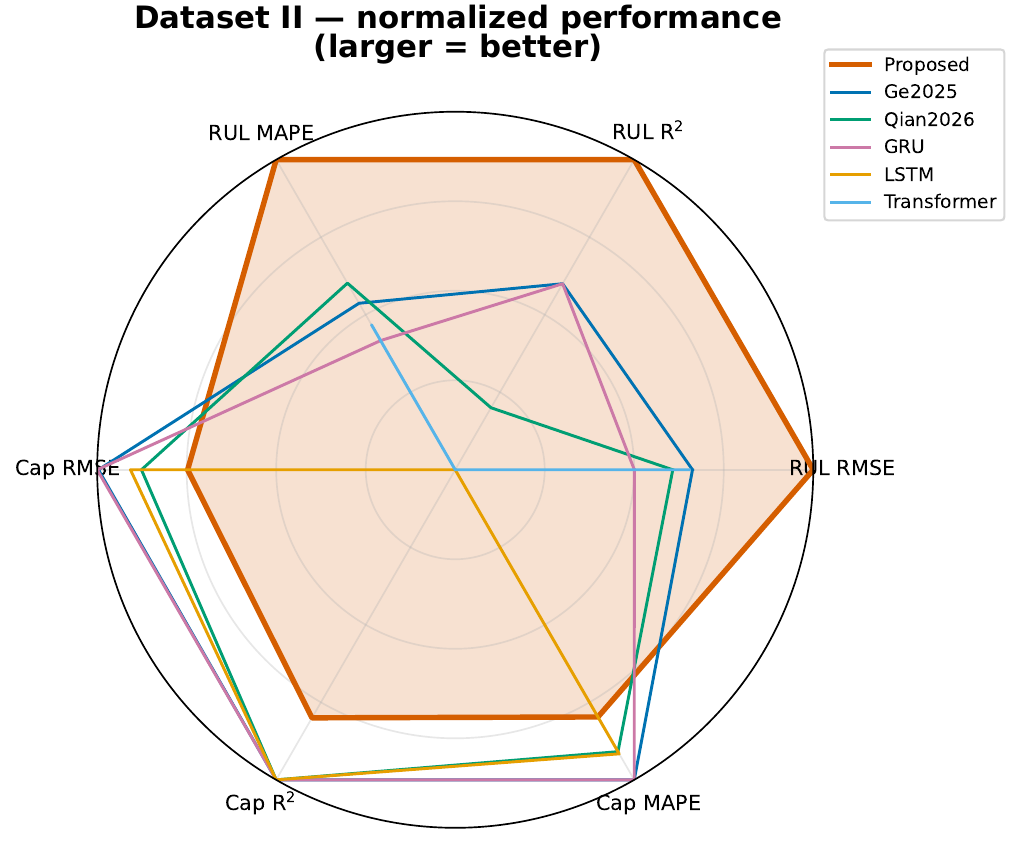}
\caption{Normalized performance profiles on (a) Dataset~I and (b) Dataset~II.}
\label{fig:radar}
\end{figure*}

The evaluation first compares the complete cross-expert framework with rerun
baselines and published reference methods on both datasets. Controlled ablations
on Dataset~I then follow the same construction sequence as the proposed method:
long-term and short-term input design, observation-window selection, Stage-I
encoder selection, Stage-II expert/backbone selection, and Stage-III fusion and
loss weighting. The reference FiLM configuration uses
$\lambda_{\mathrm{Cap}}=1$; the $\lambda_{\mathrm{Cap}}=0.5$ setting is
reported separately as an RUL-prioritized ablation. Tables~\ref{tab:sota_comparison_datasets} and~\ref{tab:main_results_verified} report means only for consistency with
published reference results that do not provide between-cell standard
deviations. Throughout this section, RUL RMSE is expressed in cycles, capacity
RMSE in mAh, and MAPE in percent; a dash denotes an unavailable result.
Boldface marks the best mean value among directly comparable configurations;
in grouped ablation tables, boldface is applied within each controlled block.

Figure~\ref{fig:radar} summarizes the six methods with complete metrics on each
dataset. Each metric is min--max normalized independently within its dataset,
with error metrics reversed so that larger values indicate better performance.
The radar plots are used only to visualize task-specific trade-offs; the
absolute metrics in Tables~\ref{tab:sota_comparison_datasets} and
\ref{tab:main_results_verified} remain the basis for comparison.

\begin{table*}[t]
\centering
\caption{Comparison with rerun baselines and published reference methods on Datasets~I and II. Boldface denotes the best available mean value for each metric within each dataset.}
\label{tab:sota_comparison_datasets}
\resizebox{\textwidth}{!}{%
\begin{tabular}{lcccccccccccc}
\toprule
\multirow{3}{*}{\textbf{Method}} &
\multicolumn{6}{c}{\textbf{Dataset I}} &
\multicolumn{6}{c}{\textbf{Dataset II}} \\
\cmidrule(lr){2-7}
\cmidrule(lr){8-13}
&
\multicolumn{3}{c}{\textbf{RUL Prediction}} &
\multicolumn{3}{c}{\textbf{Capacity Estimation}} &
\multicolumn{3}{c}{\textbf{RUL Prediction}} &
\multicolumn{3}{c}{\textbf{Capacity Estimation}} \\
\cmidrule(lr){2-4}
\cmidrule(lr){5-7}
\cmidrule(lr){8-10}
\cmidrule(lr){11-13}
&
\textbf{RMSE (cycles) $\downarrow$} &
\textbf{$R^2$ $\uparrow$} &
\textbf{MAPE (\%) $\downarrow$} &
\textbf{RMSE (mAh) $\downarrow$} &
\textbf{$R^2$ $\uparrow$} &
\textbf{MAPE (\%) $\downarrow$}  &
\textbf{RMSE (cycles) $\downarrow$} &
\textbf{$R^2$ $\uparrow$} &
\textbf{MAPE (\%) $\downarrow$} &
\textbf{RMSE (mAh) $\downarrow$} &
\textbf{$R^2$ $\uparrow$} &
\textbf{MAPE (\%) $\downarrow$} \\
\midrule
MSFEH~\cite{yu2025multi}
& 161.81 & - & \textbf{9.82} & - & - & -
& - & - & - & - & - & - \\
Structural Pruning~\cite{ge2024structural}
& - & - & - & - & - & -
& 192.17 & 0.858 & - & 6.18 & 0.995 & - \\
Ma2022~\cite{ma2022real}
& - & - & - & - & - & -
& 186 & 0.804 & 8.72 & \textbf{2.57} & 0.999 & \textbf{0.176} \\
Ge2025~\cite{ge2025deep}
& 274.71 & 0.11 & 21.67 & 13.51 & 0.87 & 1.17
& 171.82 & 0.86 & 8.15 & 3.13 & \textbf{1.00} & 0.21 \\
Qian2026~\cite{qian2026transfer}
& 290.76 & 0.02 & 23.25 & 32.91 & 0.32 & 2.76
& 173.57 & 0.84 & 8.01 & 5.13 & \textbf{1.00} & 0.34 \\
\midrule
GRU
& 252.90 & 0.30 & 18.48 & \textbf{9.39} & \textbf{0.93} & \textbf{0.59}
& 176.98 & 0.86 & 8.41 & 3.07 & \textbf{1.00} & 0.21 \\
LSTM
& 277.03 & 0.13 & 21.46 & 24.13 & 0.64 & 1.85
& 192.88 & 0.83 & 9.31 & 4.62 & \textbf{1.00} & 0.33 \\
Transformer
& 212.44 & 0.44 & 16.93 & 15.13 & 0.85 & 1.11
& 172.11 & 0.83 & 8.30 & 19.73 & 0.95 & 1.64 \\
\midrule
Proposed method
& \textbf{143.69} & \textbf{0.74} & 10.91 & 12.36 & 0.90 & 0.99
& \textbf{161.10} & \textbf{0.88} & \textbf{7.15} & 7.28 & 0.99 & 0.50 \\
\bottomrule
\end{tabular}%
}
\end{table*}

\begin{table*}[t]
\centering
\caption{Performance of baselines, standalone experts, and FiLM configurations on Dataset~I. Boldface denotes the best mean value for each metric.}
\label{tab:main_results_verified}
\footnotesize
\setlength{\tabcolsep}{5pt}
\begin{tabular}{lrrrrrr}
\toprule
& \multicolumn{3}{c}{\textbf{RUL Prediction}}
& \multicolumn{3}{c}{\textbf{Capacity Estimation}} \\
\cmidrule(lr){2-4}\cmidrule(lr){5-7}
\textbf{Method} & \textbf{RMSE (cycles) $\downarrow$} &
\textbf{$R^2$ $\uparrow$} &
\textbf{MAPE (\%) $\downarrow$} &
\textbf{RMSE (mAh) $\downarrow$} &
\textbf{$R^2$ $\uparrow$} &
\textbf{MAPE (\%) $\downarrow$} \\
\midrule
GRU         & 252.90 & 0.30 & 18.48 & \textbf{9.39}  & \textbf{0.93} & \textbf{0.59} \\
LSTM        & 277.03 & 0.13 & 21.46 & 24.13 & 0.64 & 1.85 \\
Transformer & 212.44 & 0.44 & 16.93 & 15.13 & 0.85 & 1.11 \\
RUL Expert           & 147.08 & 0.72 & 11.44 & 34.25 & 0.27 & 2.98 \\
Capacity Expert      & 151.04 & 0.70 & 11.21 & 14.41 & 0.86 & 1.10 \\
FiLM, $\lambda_{\mathrm{Cap}}=1$    & 143.69 & \textbf{0.74} & \textbf{10.91} & 12.36 & 0.90 & 0.99 \\
FiLM, $\lambda_{\mathrm{Cap}}=0.5$  & \textbf{141.64} & \textbf{0.74} & \textbf{10.91} & 15.50 & 0.84 & 1.26 \\
\bottomrule
\end{tabular}
\end{table*}

\subsection{RUL Prediction Performance}
Table~\ref{tab:sota_comparison_datasets} compares the reference FiLM model with
the rerun baselines and published methods, while
Table~\ref{tab:main_results_verified} additionally reports the two standalone
experts and the RUL-prioritized FiLM variant on Dataset~I. The reference model
achieves an RUL RMSE of 143.69 cycles, $R^2=0.74$, and MAPE of 10.91\% on
Dataset~I. Relative to the rerun GRU, LSTM, and Transformer baselines, its RMSE
is lower by 43.18\%, 48.13\%, and 32.36\%, respectively. Because these
baselines use a simpler common partial-charging input, this comparison reflects
the complete prediction pipeline rather than FiLM alone.

The standalone RUL and Capacity Experts already achieve RUL RMSEs of 147.08 and
151.04 cycles, respectively. Equal-weight FiLM further reduces these errors by
2.30\% and 4.87\%, showing that the final gain is obtained after combining two
already informative temporal representations. This result is consistent with
the proposed design: the RUL Expert represents gradual degradation over an
extended cycle history, while the Capacity Expert contributes recent-condition
information.

On Dataset~II, the reference model achieves an RUL RMSE of 161.10 cycles,
$R^2=0.88$, and MAPE of 7.15\%. It gives the lowest RUL RMSE among the listed
methods on both datasets. MSFEH nevertheless reports a lower Dataset~I RUL MAPE
(9.82\%), so the advantage is metric dependent. The smaller RMSE improvement on
Dataset~II also indicates that the magnitude of the gain depends on the data
distribution. Reducing $\lambda_{\mathrm{Cap}}$ to 0.5 further lowers Dataset~I
RUL RMSE to 141.64 cycles, but this comes at the cost of higher capacity error;
the loss-weight trade-off is examined in Section~\ref{sec:results} below.

\subsection{Capacity Estimation Performance}
The capacity results evaluate whether the joint framework preserves information
about the battery's present condition while improving RUL prediction. On
Dataset~I, the equal-weight FiLM model achieves a capacity RMSE of 12.36~mAh,
$R^2=0.90$, and MAPE of 0.99\%. This RMSE is 14.23\% lower than that of the
standalone Capacity Expert (14.41~mAh) and 63.91\% lower than that of the
standalone RUL Expert (34.25~mAh). The large difference between the two experts
is consistent with their intended roles: the Capacity Expert uses longer
40-min segments and ten consecutive cycles to emphasize recent condition,
whereas the RUL Expert uses shorter 10-min segments distributed over a 30-cycle
history.

The GRU baseline achieves a lower Dataset~I capacity RMSE of 9.39~mAh, but its
RUL RMSE is substantially higher at 252.90 cycles. Likewise, the
RUL-prioritized FiLM variant improves RUL RMSE to 141.64 cycles but increases
capacity RMSE to 15.50~mAh. Thus, the equal-weight reference is retained as the
main joint-task configuration because it improves both outputs relative to the
standalone experts rather than optimizing either task in isolation.

On Dataset~II, the reference model achieves a capacity RMSE of 7.28~mAh,
$R^2=0.99$, and MAPE of 0.50\%. Several single-task or published methods obtain
lower capacity errors, whereas the proposed method provides the lowest RUL RMSE
among the listed methods. The contribution of the framework should therefore be
interpreted as a joint RUL--capacity trade-off from partial-charging
observations, not as uniformly best capacity estimation.

% \begin{figure*}[t]
% \centering
% \includegraphics[width=0.49\linewidth]{Images/sota_combined_First_data_RUL.pdf}
% \includegraphics[width=0.49\linewidth]{Images/sota_combined_First_data_Capacity.pdf}
% \caption{RUL (left) and capacity (right) trajectories for a representative cell from Dataset~I (b3c14). Smoothed curves are shown for visualization only.}
% \label{fig:sota_combined_first}
% \end{figure*}

% \begin{figure*}[t]
% \centering
% \includegraphics[width=0.49\linewidth]{Images/sota_combined_Second_data_RUL.pdf}
% \includegraphics[width=0.49\linewidth]{Images/sota_combined_Second_data_Capacity.pdf}
% \caption{RUL (left) and capacity (right) trajectories for a representative cell from Dataset~II (10-6). Smoothed curves are shown for visualization only.}
% \label{fig:sota_combined_second}
% \end{figure*}

\begin{figure*}[t]
\centering

\includegraphics[width=0.24\textwidth]{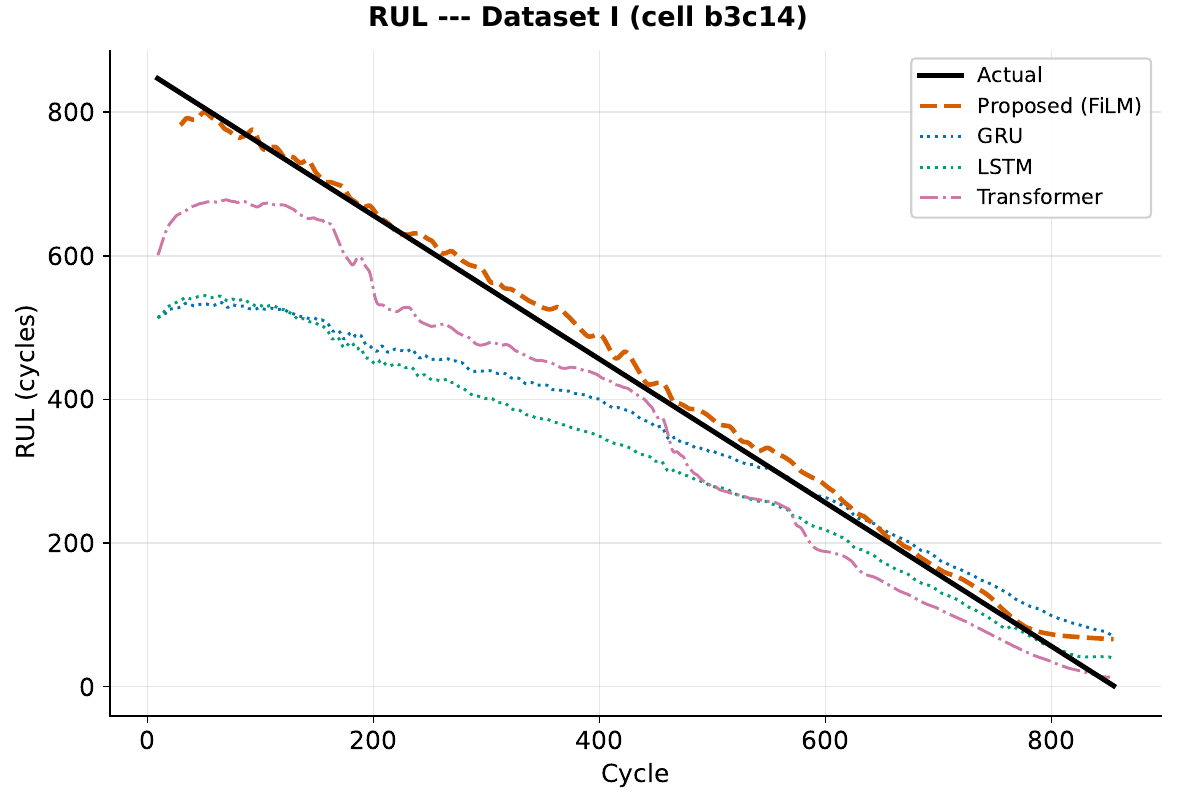}
\hfill
\includegraphics[width=0.24\textwidth]{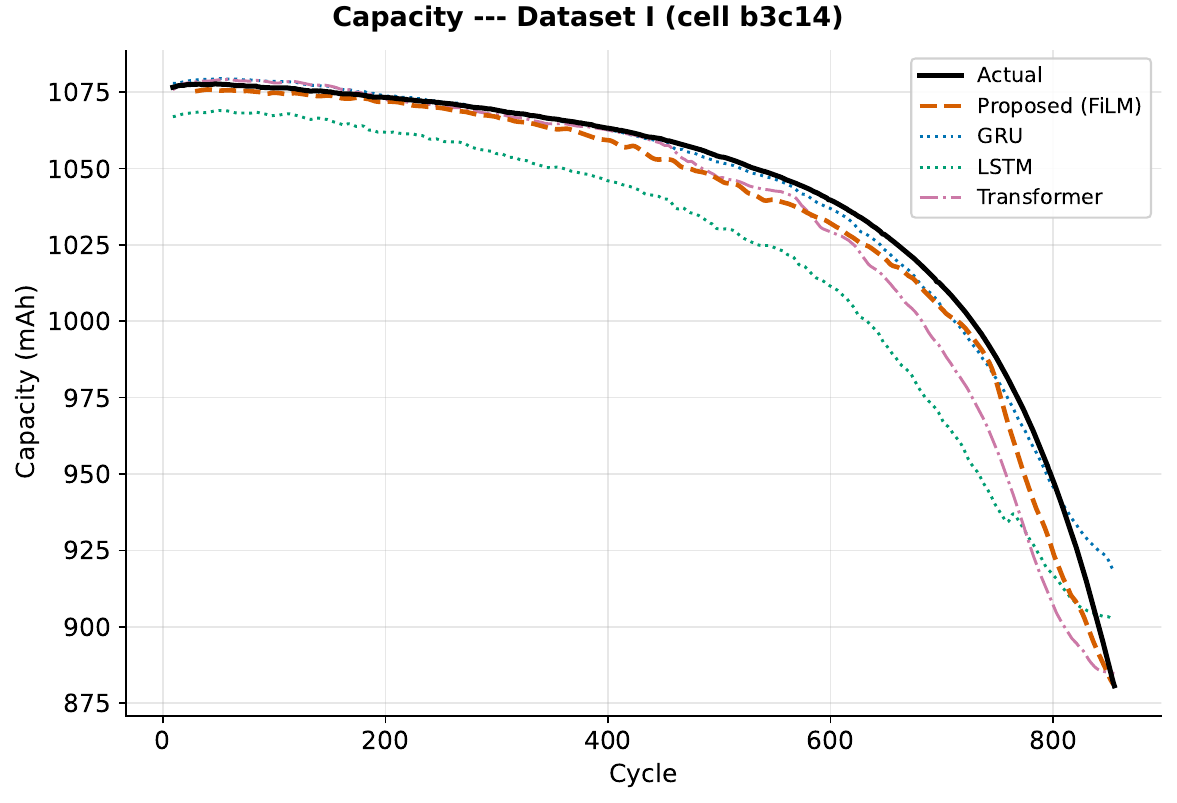}
\hfill
\includegraphics[width=0.24\textwidth]{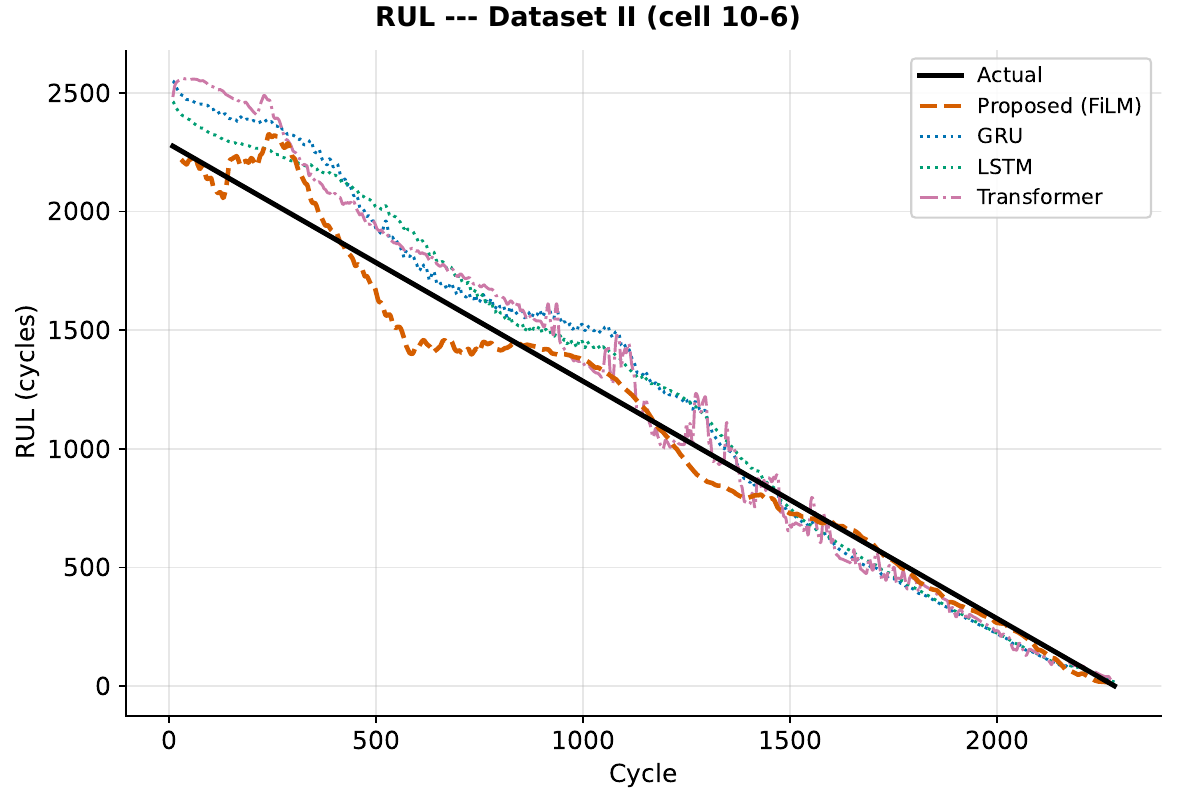}
\hfill
\includegraphics[width=0.24\textwidth]{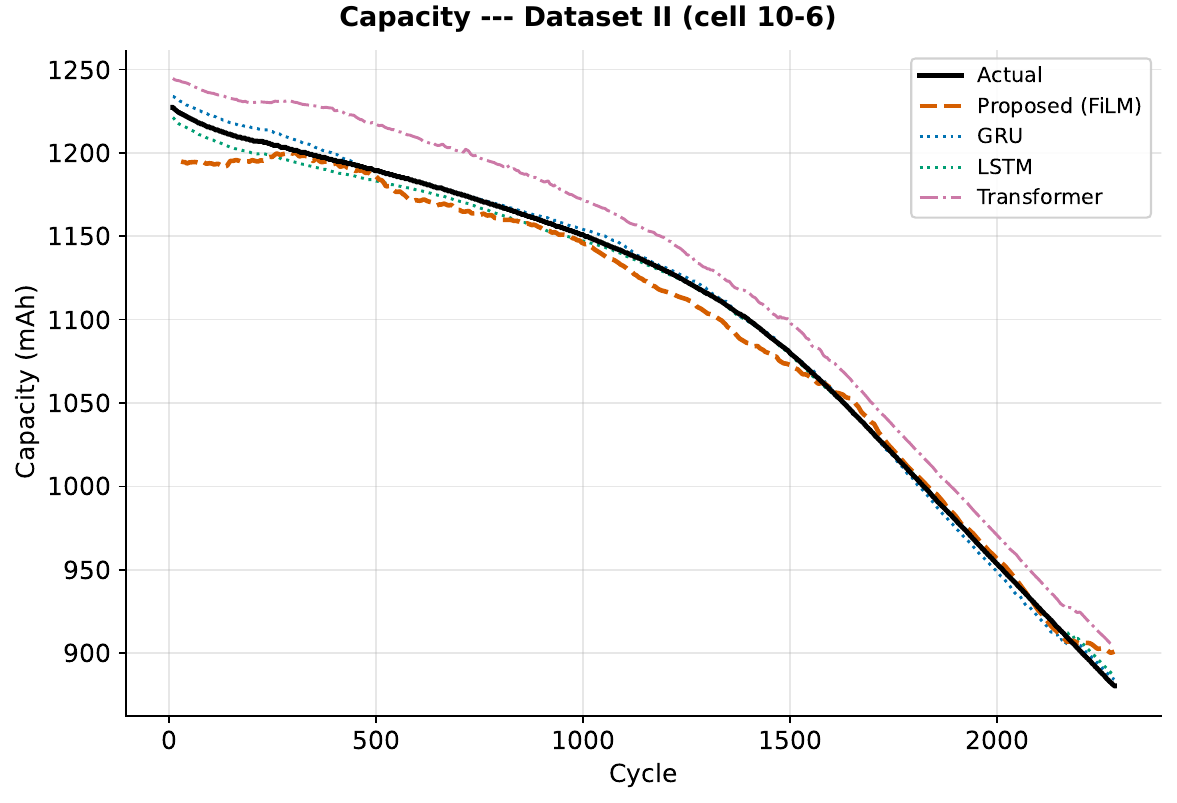}

\caption{Representative prediction trajectories: Dataset~I RUL, Dataset~I capacity, Dataset~II RUL, and Dataset~II capacity (left to right). Smoothed curves are shown for visualization only.}
\label{fig:sota_combined}
\end{figure*}

\subsection{Qualitative Prediction Trajectories}

Figure~\ref{fig:sota_combined} illustrates representative RUL and
capacity trajectories over the battery lifetime for Datasets~I and~II,
complementing the aggregate quantitative results reported above.

For Dataset~I, the proposed method follows the RUL degradation
trajectory more closely than the GRU, LSTM, and Transformer baselines
over most of the cell lifetime [Fig.~\ref{fig:sota_combined}(a)].
The capacity prediction also captures the nonlinear degradation trend
and the accelerated decline near end of life
[Fig.~\ref{fig:sota_combined}(b)], although some baselines remain
competitive for capacity estimation.

For Dataset~II, the proposed model captures the overall long-term RUL
degradation trend despite local deviations during the early and middle
aging periods [Fig.~\ref{fig:sota_combined}(c)]. The capacity prediction
similarly follows the measured degradation trajectory across the extended
cycle life [Fig.~\ref{fig:sota_combined}(d)]. Overall, these qualitative
examples support the quantitative results, showing improved long-term
RUL tracking while maintaining effective capacity estimation.

\begin{figure*}[t]
    \centering
    \includegraphics[width=\textwidth]{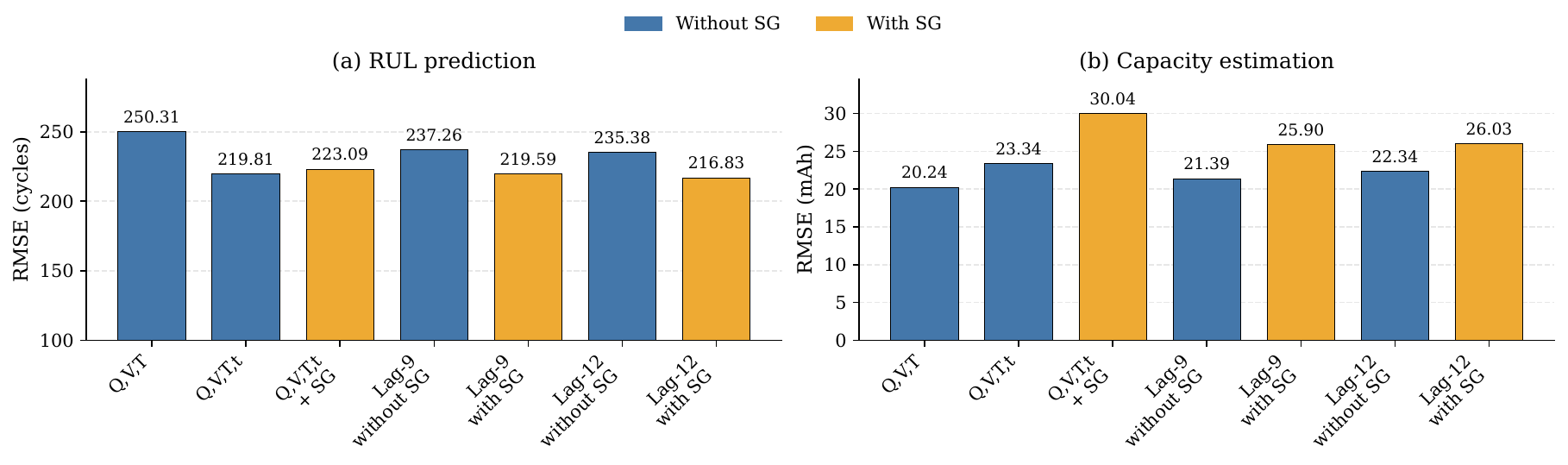}
    \caption{Effect of input features for the RUL Expert branch on (a) RUL RMSE and (b) capacity RMSE on Dataset~I.}
    \label{fig:input_feature_rmse}
\end{figure*}

\begin{figure*}[t]
    \centering
    \includegraphics[width=\textwidth]{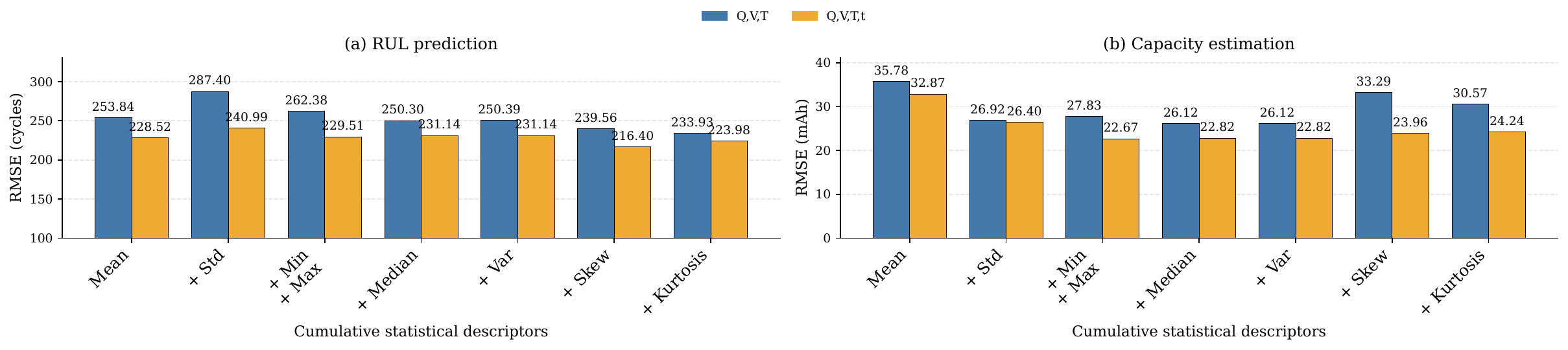}
    \caption{Effect of cumulative statistical descriptors for the Capacity Expert branch on (a) RUL RMSE and (b) capacity RMSE on Dataset~I.}
    \label{fig:statistical_descriptor_rmse}
\end{figure*}

\subsection{Ablation Studies}
All ablations use Dataset~I and are organized to mirror the proposed
three-stage framework. Representation-level experiments first isolate the
long-term RUL-Expert inputs and short-term Capacity-Expert descriptors using a
common XGBoost predictor. Observation-window and encoder studies then determine
the nominal 10-min RUL-Expert input and the Stage-I GRU encoder. The
cycle-history/backbone experiment selects the Stage-II expert architectures,
and the final fusion and task-weight experiments evaluate Stage~III. Results
from different controlled settings are not compared as if they differed by only
one factor when preprocessing or predictor architecture also changes.

\subsubsection{Input Features and Statistical Descriptors}
The first two ablations correspond directly to the branch-specific input
construction in Section~\ref{sec:proposed_method}. Figure~\ref{fig:input_feature_rmse}
examines the signal-level channels used by the RUL Expert, whereas
Fig.~\ref{fig:statistical_descriptor_rmse} examines the statistical descriptors
used by the Capacity Expert. XGBoost is kept fixed so that changes can be
attributed to the representation rather than to a neural backbone.

\begin{table*}[t]
\centering
\caption{Voltage- and time-window ablation study on Dataset~I using different input representations with a common XGBoost predictor. Boldface denotes the best mean value within each input-representation block.}
\label{tab:ablation_voltage}
\footnotesize
\setlength{\tabcolsep}{3.5pt}
\renewcommand{\arraystretch}{1.08}

\resizebox{\textwidth}{!}{%
\begin{tabular}{clcccccc}
\toprule
\multirow{2}{*}{\textbf{Input Representation}} &
\multirow{2}{*}{\textbf{Voltage / Time Window}} &
\multicolumn{3}{c}{\textbf{RUL Prediction}} &
\multicolumn{3}{c}{\textbf{Capacity Estimation}} \\
\cmidrule(lr){3-5}
\cmidrule(lr){6-8}
& &
\textbf{RMSE (cycles) $\downarrow$} &
\textbf{$R^2$ $\uparrow$} &
\textbf{MAPE (\%) $\downarrow$} &
\textbf{RMSE (mAh) $\downarrow$} &
\textbf{$R^2$ $\uparrow$} &
\textbf{MAPE (\%) $\downarrow$} \\
\midrule

% ============================================================
% Statistic Features
% ============================================================
\multirow{11}{*}{
\begin{tabular}[c]{@{}c@{}}
Statistical\\Features
\end{tabular}}
& $[3.4, 3.6]$ & 216.40 $\pm$ 142.56 & 0.44 $\pm$ 0.38 & 16.51 $\pm$ 5.90 & 23.96 $\pm$ 5.87 & 0.66 $\pm$ 0.15 & 1.70 $\pm$ 0.43 \\
& $[3.3, 3.5]$ & 229.88 $\pm$ 162.24 & 0.36 $\pm$ 0.47 & 17.57 $\pm$ 7.64 & 21.31 $\pm$ 9.30 & 0.71 $\pm$ 0.25 & 1.58 $\pm$ 0.72 \\
& $[3.2, 3.4]$ & 269.71 $\pm$ 169.62 & 0.03 $\pm$ 0.98 & 21.25 $\pm$ 10.84 & 36.46 $\pm$ 18.28 & 0.08 $\pm$ 0.91 & 2.83 $\pm$ 1.61 \\
& $[3.1, 3.3]$ & 256.43 $\pm$ 142.50 & 0.01 $\pm$ 1.49 & 20.36 $\pm$ 10.68 & 39.38 $\pm$ 18.76 & -0.06 $\pm$ 0.90 & 3.18 $\pm$ 1.74 \\
& $[3.3, 3.6]$ & 255.40 $\pm$ 158.06 & 0.23 $\pm$ 0.47 & 19.74 $\pm$ 6.68 & 26.29 $\pm$ 9.91 & 0.57 $\pm$ 0.32 & 1.90 $\pm$ 0.82 \\
& $[3.2, 3.6]$ & 226.29 $\pm$ 145.81 & 0.40 $\pm$ 0.40 & 17.14 $\pm$ 6.12 & 26.26 $\pm$ 7.66 & 0.57 $\pm$ 0.25 & 1.90 $\pm$ 0.64 \\
& $[3.1, 3.6]$ & 205.64 $\pm$ 140.99 & \textbf{0.51 $\pm$ 0.36} & 15.72 $\pm$ 5.90 & 24.69 $\pm$ 6.76 & 0.63 $\pm$ 0.20 & 1.71 $\pm$ 0.56 \\
& $V > 3.1$\,V, $(t{:}t{+}10)$ & 214.55 $\pm$ 145.24 & 0.45 $\pm$ 0.40 & 16.44 $\pm$ 5.69 & 32.41 $\pm$ 6.48 & 0.37 $\pm$ 0.32 & 2.45 $\pm$ 0.66 \\
& $V > 3.1$\,V, $(t{:}t{+}20)$ & 267.25 $\pm$ 156.63 & 0.14 $\pm$ 0.49 & 21.61 $\pm$ 7.49 & 14.74 $\pm$ 4.26 & 0.87 $\pm$ 0.09 & 1.18 $\pm$ 0.35 \\
& $V > 3.1$\,V, $(t{:}t{+}30)$ & 220.62 $\pm$ 135.48 & 0.40 $\pm$ 0.42 & 16.38 $\pm$ 6.50 & \textbf{10.86 $\pm$ 3.45} & \textbf{0.93 $\pm$ 0.05} & \textbf{0.81 $\pm$ 0.26} \\
& $V > 3.1$\,V, $(t{:}t{+}40)$ & \textbf{199.98 $\pm$ 138.06} & \textbf{0.51 $\pm$ 0.37} & \textbf{14.72 $\pm$ 6.39} & 12.77 $\pm$ 3.83 & 0.90 $\pm$ 0.07 & 0.93 $\pm$ 0.35 \\

\midrule

% ============================================================
% Partial-Charge Embedding Features
% ============================================================
\multirow{11}{*}{
\begin{tabular}[c]{@{}c@{}}
Embedding\\Features

\end{tabular}}
& $[3.4, 3.6]$ & 216.83 $\pm$ 141.44 & 0.43 $\pm$ 0.38 & 16.96 $\pm$ 6.36 & 26.03 $\pm$ 8.23 & 0.57 $\pm$ 0.29 & 1.93 $\pm$ 0.75 \\
& $[3.3, 3.5]$ & 175.66 $\pm$ 144.57 & 0.61 $\pm$ 0.38 & 12.84 $\pm$ 6.93 & 22.15 $\pm$ 10.24 & 0.68 $\pm$ 0.31 & 1.63 $\pm$ 0.90 \\
& $[3.2, 3.4]$ & 206.67 $\pm$ 123.98 & 0.37 $\pm$ 0.66 & 16.46 $\pm$ 7.95 & 31.84 $\pm$ 13.22 & 0.34 $\pm$ 0.48 & 2.37 $\pm$ 0.96 \\
& $[3.1, 3.3]$ & 193.84 $\pm$ 133.89 & 0.28 $\pm$ 1.42 & 15.85 $\pm$ 11.79 & 33.74 $\pm$ 17.58 & 0.20 $\pm$ 0.75 & 2.59 $\pm$ 1.49 \\
& $[3.3, 3.6]$ & 232.85 $\pm$ 135.60 & 0.33 $\pm$ 0.40 & 18.28 $\pm$ 6.21 & 27.94 $\pm$ 12.77 & 0.48 $\pm$ 0.44 & 2.05 $\pm$ 1.03 \\
& $[3.2, 3.6]$ & 226.82 $\pm$ 139.97 & 0.38 $\pm$ 0.39 & 17.64 $\pm$ 6.00 & 28.10 $\pm$ 13.75 & 0.47 $\pm$ 0.49 & 2.10 $\pm$ 1.14 \\
& $[3.1, 3.6]$ & 228.23 $\pm$ 145.49 & 0.37 $\pm$ 0.41 & 17.67 $\pm$ 6.28 & 28.69 $\pm$ 12.13 & 0.47 $\pm$ 0.39 & 2.14 $\pm$ 0.96 \\
& $V > 3.1$\,V, $(t{:}t{+}10)$ & \textbf{159.38 $\pm$ 126.86} & \textbf{0.68 $\pm$ 0.29} & \textbf{12.21 $\pm$ 6.35} & 36.66 $\pm$ 7.78 & 0.19 $\pm$ 0.41 & 3.07 $\pm$ 0.77 \\
& $V > 3.1$\,V, $(t{:}t{+}20)$ & 316.30 $\pm$ 155.38 & -0.17 $\pm$ 0.48 & 25.93 $\pm$ 5.75 & 21.13 $\pm$ 8.75 & 0.72 $\pm$ 0.28 & 1.67 $\pm$ 0.82 \\
& $V > 3.1$\,V, $(t{:}t{+}30)$ & 286.84 $\pm$ 145.60 & 0.03 $\pm$ 0.42 & 23.36 $\pm$ 5.68 & \textbf{19.59 $\pm$ 6.12} & \textbf{0.77 $\pm$ 0.16} & \textbf{1.50 $\pm$ 0.59} \\
& $V > 3.1$\,V, $(t{:}t{+}40)$ & 285.96 $\pm$ 144.86 & 0.05 $\pm$ 0.40 & 23.29 $\pm$ 5.49 & 23.77 $\pm$ 7.85 & 0.65 $\pm$ 0.26 & 1.90 $\pm$ 0.72 \\

\midrule

% ============================================================
% Interpolated Features
% ============================================================
\multirow{11}{*}{
\begin{tabular}[c]{@{}c@{}}
Interpolated\\Features
\end{tabular}}
& $[3.4, 3.6]$ & 241.94 $\pm$ 147.19 & 0.32 $\pm$ 0.38 & 18.63 $\pm$ 5.47 & 28.79 $\pm$ 7.76 & 0.49 $\pm$ 0.31 & 2.10 $\pm$ 0.62 \\
& $[3.3, 3.5]$ & \textbf{162.27 $\pm$ 149.71} & \textbf{0.69 $\pm$ 0.35} & \textbf{11.29 $\pm$ 6.73} & 21.32 $\pm$ 9.54 & 0.71 $\pm$ 0.23 & 1.47 $\pm$ 0.69 \\
& $[3.2, 3.4]$ & 287.75 $\pm$ 182.36 & -0.77 $\pm$ 3.93 & 24.14 $\pm$ 20.29 & 35.15 $\pm$ 13.56 & 0.21 $\pm$ 0.54 & 2.61 $\pm$ 1.13 \\
& $[3.1, 3.3]$ & 256.54 $\pm$ 162.09 & -0.42 $\pm$ 3.13 & 21.85 $\pm$ 18.32 & 39.04 $\pm$ 17.09 & -0.01 $\pm$ 0.78 & 3.08 $\pm$ 1.53 \\
& $[3.3, 3.6]$ & 261.79 $\pm$ 134.44 & 0.20 $\pm$ 0.36 & 20.19 $\pm$ 4.75 & 29.14 $\pm$ 9.32 & 0.47 $\pm$ 0.34 & 2.11 $\pm$ 0.84 \\
& $[3.2, 3.6]$ & 240.85 $\pm$ 127.05 & 0.31 $\pm$ 0.34 & 18.26 $\pm$ 4.94 & 31.12 $\pm$ 9.69 & 0.39 $\pm$ 0.39 & 2.30 $\pm$ 0.89 \\
& $[3.1, 3.6]$ & 245.70 $\pm$ 133.04 & 0.29 $\pm$ 0.35 & 18.76 $\pm$ 4.73 & 32.55 $\pm$ 10.02 & 0.33 $\pm$ 0.42 & 2.45 $\pm$ 0.88 \\
& $V > 3.1$\,V, $(t{:}t{+}10)$ & 237.17 $\pm$ 153.88 & 0.35 $\pm$ 0.44 & 17.92 $\pm$ 5.89 & 34.49 $\pm$ 6.70 & 0.29 $\pm$ 0.34 & 2.57 $\pm$ 0.71 \\
& $V > 3.1$\,V, $(t{:}t{+}20)$ & 306.44 $\pm$ 157.22 & -0.07 $\pm$ 0.47 & 24.48 $\pm$ 5.84 & \textbf{17.37 $\pm$ 4.69} & \textbf{0.81 $\pm$ 0.10} & \textbf{1.44 $\pm$ 0.45} \\
& $V > 3.1$\,V, $(t{:}t{+}30)$ & 316.40 $\pm$ 165.56 & -0.15 $\pm$ 0.54 & 24.34 $\pm$ 6.61 & 27.11 $\pm$ 6.76 & 0.56 $\pm$ 0.24 & 2.36 $\pm$ 0.67 \\
& $V > 3.1$\,V, $(t{:}t{+}40)$ & 342.38 $\pm$ 162.34 & -0.38 $\pm$ 0.62 & 27.28 $\pm$ 6.75 & 41.84 $\pm$ 5.63 & 0.01 $\pm$ 0.25 & 3.51 $\pm$ 0.53 \\

\bottomrule
\end{tabular}
}
\end{table*}

For the RUL Expert, adding elapsed time to $Q,V,T$ reduces RUL RMSE from
250.31 to 219.81 cycles. Savitzky--Golay filtering is most useful when combined
with inter-cycle differences: it reduces RUL RMSE from 237.26 to 219.59 cycles
for lag-9 and from 235.38 to 216.83 cycles for lag-12. The filtered lag-12
representation gives the lowest RUL RMSE among the evaluated inputs, supporting
the long-term branch design in which the filtered $Q,V,T,t$ channels are
augmented with lag-12 differences. The corresponding capacity errors increase,
which is consistent with selecting this representation for the RUL-oriented
long-term view rather than for short-term capacity estimation.

For the Capacity Expert, the descriptors are accumulated over the four
$Q,V,T,t$ channels. The seven-statistic set used in the proposed method---mean,
standard deviation, minimum, maximum, median, variance, and skewness---reduces
RUL RMSE from 228.52 to 216.40 cycles and gives a capacity RMSE of
23.96~mAh. Four descriptors (mean, standard deviation, minimum, and maximum)
produce a lower capacity RMSE of 22.67~mAh, while adding kurtosis increases RUL
RMSE to 223.98 cycles. The seven-statistic set is therefore retained as the
28-dimensional short-term representation as a joint-task choice rather than as
the capacity-only optimum.

\subsubsection{Voltage and Time-Window}
Table~\ref{tab:ablation_voltage} evaluates the effect of the observed charging region under statistical, embedding, and interpolated input representations. The optimal window depends strongly on both the representation and prediction objective. For the statistical features, the 40-min segment achieves the lowest RUL RMSE of 199.98 cycles, whereas the 30-min segment provides the lowest capacity RMSE of 10.86~mAh. For the partial-charge embeddings, the 10-min segment performs best for RUL prediction, achieving an RMSE of 159.38 cycles, while extending the observation to 30~min reduces the capacity RMSE to 19.59~mAh. The interpolated representation shows a similar task-dependent behavior: the $[3.3,3.5]$~V interval gives the lowest RUL RMSE of 162.27 cycles, whereas the 20-min segment gives the lowest capacity RMSE of 17.37~mAh. These results show that increasing the voltage range or observation duration does not consistently improve both tasks and that the charging window should be selected jointly with the feature representation.

Based on this trade-off, the RUL Expert uses the 10-min partial-charge embedding representation because it provides the strongest RUL performance among the embedding configurations. In contrast, the Capacity Expert uses the longer 40-min statistical representation. Although the 30-min statistical window gives the lowest standalone capacity RMSE, the 40-min window substantially improves RUL prediction while retaining a low capacity error of 12.77~mAh. This choice is therefore consistent with the joint prediction objective and the overall design preference toward RUL accuracy rather than optimizing capacity estimation independently.

\subsubsection{Autoencoder Architecture}
Table~\ref{tab:rd_ae} compares jointly supervised autoencoder architectures using the same XGBoost predictor. GRU achieves the lowest RUL RMSE and MAPE, with 156.31 cycles and 11.88\%, respectively, while LSTM provides a slightly higher $R^2$ of 0.70. In contrast, H3MAE gives the strongest capacity-estimation results, with an RMSE of 27.87~mAh, $R^2$ of 0.52, and MAPE of 2.01\%. These results further demonstrate that the representations most effective for RUL prediction are not necessarily optimal for capacity estimation.

Because the pretrained encoder is transferred specifically to the long-term RUL Expert and the proposed framework prioritizes RUL prediction while maintaining joint supervision from both targets, GRU is selected as the Stage-I encoder. Its superior RUL RMSE and MAPE provide the strongest degradation-oriented representation for the subsequent long-term expert, while capacity information is still incorporated through the auxiliary capacity loss during supervised autoencoder pretraining.

\begin{table*}[t]
\centering
\caption{Comparison of jointly supervised autoencoder architectures with XGBoost regression on Dataset~I. Boldface denotes the best mean value for each metric.}
\label{tab:rd_ae}
\footnotesize
\setlength{\tabcolsep}{4pt}
\renewcommand{\arraystretch}{1.08}
\resizebox{0.8\textwidth}{!}{%
\begin{tabular}{lcccccc}
\toprule
\multirow{2}{*}{\textbf{Configuration}} & \multicolumn{3}{c}{\textbf{RUL Prediction}} & \multicolumn{3}{c}{\textbf{Capacity Estimation}}\\
\cmidrule(lr){2-4}\cmidrule(lr){5-7}
& \textbf{RMSE (cycles) $\downarrow$} &
\textbf{$R^2$ $\uparrow$} &
\textbf{MAPE (\%) $\downarrow$} &
\textbf{RMSE (mAh) $\downarrow$} &
\textbf{$R^2$ $\uparrow$} &
\textbf{MAPE (\%) $\downarrow$} \\
\midrule
LSTM &  157.87 $\pm$ 130.12 & \textbf{0.70 $\pm$ 0.28} & 11.91 $\pm$ 5.62 & 32.64 $\pm$ 7.76 & 0.36 $\pm$ 0.36 & 2.73 $\pm$ 0.74 \\
GRU &  \textbf{156.31 $\pm$ 120.96} & 0.69 $\pm$ 0.27 & \textbf{11.88 $\pm$ 5.96} & 34.46 $\pm$ 7.19 & 0.30 $\pm$ 0.33 & 2.77 $\pm$ 0.70 \\
Ti-MAE~\cite{li2023ti} &  195.36 $\pm$ 144.24 & 0.55 $\pm$ 0.36 & 14.91 $\pm$ 6.09 & 29.02 $\pm$ 10.48 & 0.46 $\pm$ 0.40 & 2.28 $\pm$ 0.95 \\
TS-MAE~\cite{liu2025ts} &  208.55 $\pm$ 147.32 & 0.47 $\pm$ 0.41 & 15.87 $\pm$ 6.62 & 31.27 $\pm$ 10.18 & 0.39 $\pm$ 0.40 & 2.33 $\pm$ 0.97 \\
H3MAE~\cite{yuan2025hierarchical} &  213.76 $\pm$ 149.45 & 0.45 $\pm$ 0.42 & 16.59 $\pm$ 6.74 & \textbf{27.87 $\pm$ 8.53} & \textbf{0.52 $\pm$ 0.31} & \textbf{2.01 $\pm$ 0.80} \\
TimeMAE~\cite{cheng2026timemae} &  230.42 $\pm$ 145.57 & 0.36 $\pm$ 0.46 & 17.56 $\pm$ 6.68 & 31.85 $\pm$ 10.61 & 0.36 $\pm$ 0.41 & 2.32 $\pm$ 0.98 \\
\bottomrule
\end{tabular}%
}
\end{table*}

\subsubsection{Cycle History and Prediction Backbone}
Table~\ref{tab:ablation_prediction} determines the Stage-II expert structures by
jointly varying temporal coverage, input representation, and prediction
backbone. The extended setting samples ten cycles at stride three from a
30-cycle window; the recent setting uses ten consecutive cycles at stride one.
Thus, both contain ten cycle-level inputs, but the former covers gradual
long-term degradation whereas the latter emphasizes the current degradation
state. Statistical features use the selected nominal 40-min segments, while
embedding features use nominal 10-min segments encoded by the pretrained GRU.

\begin{table*}[t]
\centering
\caption{Prediction-module ablation study on Dataset~I under different cycle-sampling strategies, input representations, and prediction backbones. Boldface denotes the best mean value within each cycle-sampling/input-representation block.}
\label{tab:ablation_prediction}
\footnotesize
\setlength{\tabcolsep}{3pt}
\renewcommand{\arraystretch}{1.05}

\resizebox{\textwidth}{!}{%
\begin{tabular}{cclcccccc}
\toprule

\multirow{2}{*}{\textbf{Cycle Sampling}} &
\multirow{2}{*}{\textbf{Input Representation}} &
\multirow{2}{*}{\textbf{Backbone}} &
\multicolumn{3}{c}{\textbf{RUL Prediction}} &
\multicolumn{3}{c}{\textbf{Capacity Estimation}} \\

\cmidrule(lr){4-6}
\cmidrule(lr){7-9}

& & &
\textbf{RMSE (cycles) $\downarrow$} &
\textbf{$R^2$ $\uparrow$} &
\textbf{MAPE (\%) $\downarrow$} &
\textbf{RMSE (mAh) $\downarrow$} &
\textbf{$R^2$ $\uparrow$} &
\textbf{MAPE (\%) $\downarrow$} \\

\midrule

% ============================================================
% 30-cycle window - Statistic Features
% ============================================================
\multirow{28}{*}{
\begin{tabular}[c]{@{}c@{}}
30-cycle window\\
10 uniformly sampled cycles\\
(step = 3)
\end{tabular}}

&
\multirow{14}{*}{
\begin{tabular}[c]{@{}c@{}}
Statistical\\Features
\end{tabular}}

& 1D CNN               & 229.30 $\pm$ 141.97 & 0.36 $\pm$ 0.40 & 18.53 $\pm$ 6.70 & 27.78 $\pm$ 7.78 & 0.54 $\pm$ 0.30 & 1.88 $\pm$ 0.73 \\
& & 2D CNN               & 168.62 $\pm$ 117.76 & 0.60 $\pm$ 0.38 & 12.96 $\pm$ 6.92 & 21.46 $\pm$ 10.40 & 0.68 $\pm$ 0.30 & 1.81 $\pm$ 0.99 \\
& & LSTM                 & 254.76 $\pm$ 158.96 & 0.19 $\pm$ 0.54 & 21.19 $\pm$ 8.44 & 42.50 $\pm$ 10.42 & -0.06 $\pm$ 0.56 & 3.26 $\pm$ 0.96 \\
& & GRU                  & 232.09 $\pm$ 142.38 & 0.32 $\pm$ 0.47 & 19.55 $\pm$ 8.19 & 50.73 $\pm$ 14.20 & -0.55 $\pm$ 0.87 & 3.73 $\pm$ 1.19 \\
& & Transformer          & 264.13 $\pm$ 153.24 & 0.16 $\pm$ 0.48 & 21.48 $\pm$ 6.97 & 25.72 $\pm$ 6.75 & 0.61 $\pm$ 0.20 & 1.54 $\pm$ 0.66 \\
& & Mamba                & 292.75 $\pm$ 160.49 & -0.02 $\pm$ 0.52 & 24.38 $\pm$ 6.96 & 33.21 $\pm$ 7.62 & 0.35 $\pm$ 0.31 & 2.32 $\pm$ 0.66 \\
& & 1D CNN + LSTM        & 256.46 $\pm$ 157.64 & 0.20 $\pm$ 0.49 & 20.87 $\pm$ 7.57 & 32.70 $\pm$ 8.16 & 0.38 $\pm$ 0.33 & 2.13 $\pm$ 0.80 \\
& & 1D CNN + GRU         & 256.83 $\pm$ 153.31 & 0.20 $\pm$ 0.48 & 20.94 $\pm$ 7.35 & 33.45 $\pm$ 7.17 & 0.35 $\pm$ 0.28 & 2.37 $\pm$ 0.70 \\
& & 1D CNN + Transformer & 269.99 $\pm$ 156.72 & 0.12 $\pm$ 0.51 & 22.24 $\pm$ 7.43 & 35.54 $\pm$ 10.91 & 0.25 $\pm$ 0.50 & 2.26 $\pm$ 1.07 \\
& & 1D CNN + Mamba       & 259.35 $\pm$ 151.85 & 0.19 $\pm$ 0.46 & 21.02 $\pm$ 7.01 & 27.77 $\pm$ 8.30 & 0.54 $\pm$ 0.29 & 2.09 $\pm$ 0.73 \\
& & 2D CNN + LSTM        & \textbf{153.47 $\pm$ 113.44} & \textbf{0.70 $\pm$ 0.27} & \textbf{11.52 $\pm$ 5.67} & 22.18 $\pm$ 6.28 & 0.70 $\pm$ 0.17 & 1.48 $\pm$ 0.68 \\
& & 2D CNN + GRU         & 155.17 $\pm$ 113.80 & 0.69 $\pm$ 0.26 & 11.57 $\pm$ 5.46 & 23.65 $\pm$ 8.03 & 0.65 $\pm$ 0.24 & 1.70 $\pm$ 0.79 \\
& & 2D CNN + Transformer & 154.43 $\pm$ 115.44 & 0.66 $\pm$ 0.40 & 12.07 $\pm$ 7.03 & \textbf{14.27 $\pm$ 5.35} & \textbf{0.87 $\pm$ 0.11} & \textbf{1.09 $\pm$ 0.51} \\
& & 2D CNN + Mamba       & 168.57 $\pm$ 122.63 & 0.64 $\pm$ 0.30 & 12.87 $\pm$ 6.13 & 23.51 $\pm$ 8.32 & 0.65 $\pm$ 0.25 & 1.62 $\pm$ 0.79 \\

\cmidrule(lr){2-9}

% ============================================================
% 30-cycle window - Embedding Features
% ============================================================
&
\multirow{14}{*}{
\begin{tabular}[c]{@{}c@{}}
Embedding\\Features
\end{tabular}}

& 1D CNN               & 164.82 $\pm$ 129.34 & 0.61 $\pm$ 0.39 & 13.44 $\pm$ 8.27 & \textbf{23.78 $\pm$ 8.69} & \textbf{0.64 $\pm$ 0.28} & \textbf{1.73 $\pm$ 0.89} \\
& & 2D CNN               & 165.92 $\pm$ 134.24 & 0.63 $\pm$ 0.36 & 12.93 $\pm$ 7.72 & 30.94 $\pm$ 10.28 & 0.40 $\pm$ 0.43 & 2.69 $\pm$ 1.00 \\
& & LSTM                 & 164.22 $\pm$ 133.49 & 0.59 $\pm$ 0.50 & 13.46 $\pm$ 9.53 & 32.96 $\pm$ 9.00 & 0.33 $\pm$ 0.42 & 2.88 $\pm$ 0.87 \\
& & GRU                  & 159.08 $\pm$ 135.97 & 0.63 $\pm$ 0.43 & 12.80 $\pm$ 8.68 & 29.04 $\pm$ 7.69 & 0.48 $\pm$ 0.30 & 2.39 $\pm$ 0.76 \\
& & Transformer          & 159.03 $\pm$ 124.58 & 0.65 $\pm$ 0.33 & 12.82 $\pm$ 7.57 & 49.42 $\pm$ 8.47 & -0.44 $\pm$ 0.60 & 4.49 $\pm$ 0.79 \\
& & Mamba                & 161.62 $\pm$ 124.14 & 0.67 $\pm$ 0.27 & 12.73 $\pm$ 6.18 & 30.53 $\pm$ 8.53 & 0.43 $\pm$ 0.37 & 2.50 $\pm$ 0.87 \\
& & 1D CNN + LSTM        & 173.12 $\pm$ 128.26 & 0.61 $\pm$ 0.34 & 14.07 $\pm$ 7.15 & 32.61 $\pm$ 8.22 & 0.35 $\pm$ 0.39 & 2.77 $\pm$ 0.84 \\
& & 1D CNN + GRU         & 165.31 $\pm$ 128.12 & 0.62 $\pm$ 0.37 & 13.18 $\pm$ 7.85 & 33.60 $\pm$ 8.49 & 0.31 $\pm$ 0.39 & 2.90 $\pm$ 0.81 \\
& & 1D CNN + Transformer & 164.45 $\pm$ 127.81 & 0.64 $\pm$ 0.33 & 13.07 $\pm$ 7.27 & 44.91 $\pm$ 8.21 & -0.20 $\pm$ 0.55 & 4.02 $\pm$ 0.79 \\
& & 1D CNN + Mamba       & 163.61 $\pm$ 121.35 & 0.66 $\pm$ 0.28 & 12.86 $\pm$ 5.93 & 31.22 $\pm$ 8.27 & 0.40 $\pm$ 0.36 & 2.64 $\pm$ 0.87 \\
& & 2D CNN + LSTM        & 162.81 $\pm$ 131.40 & 0.64 $\pm$ 0.35 & 13.21 $\pm$ 7.45 & 30.73 $\pm$ 9.67 & 0.41 $\pm$ 0.39 & 2.53 $\pm$ 0.87 \\
& & 2D CNN + GRU         & \textbf{147.08 $\pm$ 121.88} & \textbf{0.72 $\pm$ 0.27} & \textbf{11.44 $\pm$ 6.20} & 34.25 $\pm$ 9.19 & 0.27 $\pm$ 0.45 & 2.98 $\pm$ 0.85 \\
& & 2D CNN + Transformer & 173.48 $\pm$ 125.65 & 0.60 $\pm$ 0.35 & 14.19 $\pm$ 7.39 & 43.42 $\pm$ 9.29 & -0.12 $\pm$ 0.55 & 3.84 $\pm$ 0.83 \\
& & 2D CNN + Mamba       & 167.28 $\pm$ 132.85 & 0.65 $\pm$ 0.31 & 13.11 $\pm$ 6.42 & 32.68 $\pm$ 8.32 & 0.34 $\pm$ 0.39 & 2.81 $\pm$ 0.80 \\

\midrule

% ============================================================
% 10-cycle window - Statistic Features
% ============================================================
\multirow{28}{*}{
\begin{tabular}[c]{@{}c@{}}
10-cycle window\\
10 consecutive cycles\\
(step = 1)
\end{tabular}}

&
\multirow{14}{*}{
\begin{tabular}[c]{@{}c@{}}
Statistical\\Features
\end{tabular}}

& 1D CNN               & 252.00 $\pm$ 145.98 & 0.28 $\pm$ 0.41 & 19.64 $\pm$ 6.08 & 24.63 $\pm$ 6.32 & 0.63 $\pm$ 0.19 & 2.06 $\pm$ 0.56 \\
& & 2D CNN               & 174.16 $\pm$ 114.40 & 0.60 $\pm$ 0.33 & 13.10 $\pm$ 6.39 & 21.61 $\pm$ 10.81 & 0.67 $\pm$ 0.32 & 1.80 $\pm$ 1.02 \\
& & LSTM                 & 282.48 $\pm$ 165.72 & 0.07 $\pm$ 0.56 & 22.77 $\pm$ 7.85 & 43.27 $\pm$ 7.94 & -0.08 $\pm$ 0.40 & 3.49 $\pm$ 0.77 \\
& & GRU                  & 224.75 $\pm$ 137.11 & 0.39 $\pm$ 0.42 & 18.29 $\pm$ 7.66 & 50.92 $\pm$ 13.13 & -0.55 $\pm$ 0.80 & 3.63 $\pm$ 1.13 \\
& & Transformer          & 280.56 $\pm$ 153.59 & 0.10 $\pm$ 0.47 & 22.48 $\pm$ 6.47 & 36.94 $\pm$ 8.09 & 0.22 $\pm$ 0.31 & 2.27 $\pm$ 0.81 \\
& & Mamba                & 282.61 $\pm$ 154.53 & 0.10 $\pm$ 0.45 & 22.60 $\pm$ 6.18 & 28.13 $\pm$ 6.48 & 0.54 $\pm$ 0.22 & 1.83 $\pm$ 0.47 \\
& & 1D CNN + LSTM        & 252.63 $\pm$ 153.53 & 0.25 $\pm$ 0.47 & 20.24 $\pm$ 7.44 & 28.55 $\pm$ 7.82 & 0.52 $\pm$ 0.27 & 1.79 $\pm$ 0.78 \\
& & 1D CNN + GRU         & 270.67 $\pm$ 151.88 & 0.15 $\pm$ 0.47 & 21.91 $\pm$ 6.98 & 33.99 $\pm$ 6.95 & 0.33 $\pm$ 0.28 & 2.39 $\pm$ 0.71 \\
& & 1D CNN + Transformer & 289.18 $\pm$ 158.28 & 0.03 $\pm$ 0.52 & 23.67 $\pm$ 7.21 & 33.87 $\pm$ 9.44 & 0.33 $\pm$ 0.39 & 1.97 $\pm$ 0.96 \\
& & 1D CNN + Mamba       & 270.74 $\pm$ 151.92 & 0.16 $\pm$ 0.45 & 21.57 $\pm$ 6.60 & 31.94 $\pm$ 6.28 & 0.41 $\pm$ 0.23 & 2.28 $\pm$ 0.63 \\
& & 2D CNN + LSTM        & 152.26 $\pm$ 110.38 & 0.70 $\pm$ 0.29 & 11.33 $\pm$ 5.97 & 18.29 $\pm$ 5.14 & 0.80 $\pm$ 0.13 & 1.22 $\pm$ 0.55 \\
& & 2D CNN + GRU         & 154.69 $\pm$ 113.42 & \textbf{0.71 $\pm$ 0.26} & 11.07 $\pm$ 5.42 & 17.41 $\pm$ 6.70 & 0.80 $\pm$ 0.16 & 1.26 $\pm$ 0.65 \\
& & 2D CNN + Transformer & \textbf{151.04 $\pm$ 109.64} & 0.70 $\pm$ 0.30 & 11.21 $\pm$ 5.94 & \textbf{14.41 $\pm$ 6.28} & \textbf{0.86 $\pm$ 0.13} & \textbf{1.10 $\pm$ 0.62} \\
& & 2D CNN + Mamba       & 153.35 $\pm$ 108.58 & \textbf{0.71 $\pm$ 0.24} & \textbf{11.05 $\pm$ 5.17} & 20.68 $\pm$ 7.78 & 0.72 $\pm$ 0.22 & 1.41 $\pm$ 0.73 \\

\cmidrule(lr){2-9}

% ============================================================
% 10-cycle window - Embedding Features
% ============================================================
&
\multirow{14}{*}{
\begin{tabular}[c]{@{}c@{}}
Embedding\\Features
\end{tabular}}

& 1D CNN               & 175.55 $\pm$ 128.72 & 0.62 $\pm$ 0.31 & 13.66 $\pm$ 6.57 & \textbf{25.53 $\pm$ 7.93} & \textbf{0.60 $\pm$ 0.25} & \textbf{1.91 $\pm$ 0.81} \\
& & 2D CNN               & 177.55 $\pm$ 137.34 & 0.61 $\pm$ 0.36 & 14.03 $\pm$ 7.36 & 39.93 $\pm$ 9.75 & 0.03 $\pm$ 0.53 & 3.36 $\pm$ 1.03 \\
& & LSTM                 & 165.74 $\pm$ 126.14 & 0.64 $\pm$ 0.33 & 13.06 $\pm$ 7.36 & 31.63 $\pm$ 8.70 & 0.38 $\pm$ 0.38 & 2.70 $\pm$ 0.83 \\
& & GRU                  & 169.88 $\pm$ 130.22 & 0.64 $\pm$ 0.31 & 13.09 $\pm$ 6.71 & 33.58 $\pm$ 8.92 & 0.31 $\pm$ 0.41 & 2.88 $\pm$ 0.86 \\
& & Transformer          & 170.05 $\pm$ 128.06 & 0.65 $\pm$ 0.29 & 13.05 $\pm$ 5.95 & 45.58 $\pm$ 8.54 & -0.23 $\pm$ 0.56 & 4.09 $\pm$ 0.81 \\
& & Mamba                & 169.05 $\pm$ 124.18 & 0.65 $\pm$ 0.29 & 13.03 $\pm$ 6.50 & 29.79 $\pm$ 8.46 & 0.45 $\pm$ 0.35 & 2.48 $\pm$ 0.84 \\
& & 1D CNN + LSTM        & \textbf{157.60 $\pm$ 133.26} & \textbf{0.68 $\pm$ 0.33} & \textbf{12.33 $\pm$ 7.16} & 36.73 $\pm$ 8.23 & 0.19 $\pm$ 0.42 & 3.10 $\pm$ 0.81 \\
& & 1D CNN + GRU         & 162.19 $\pm$ 123.59 & 0.66 $\pm$ 0.31 & 12.58 $\pm$ 7.14 & 33.79 $\pm$ 7.47 & 0.31 $\pm$ 0.37 & 2.91 $\pm$ 0.72 \\
& & 1D CNN + Transformer & 171.53 $\pm$ 128.20 & 0.63 $\pm$ 0.32 & 13.36 $\pm$ 6.81 & 50.46 $\pm$ 8.38 & -0.50 $\pm$ 0.64 & 4.57 $\pm$ 0.81 \\
& & 1D CNN + Mamba       & 167.47 $\pm$ 126.38 & 0.65 $\pm$ 0.30 & 12.80 $\pm$ 6.73 & 34.90 $\pm$ 8.52 & 0.26 $\pm$ 0.41 & 3.05 $\pm$ 0.84 \\
& & 2D CNN + LSTM        & 169.52 $\pm$ 137.17 & 0.59 $\pm$ 0.46 & 13.74 $\pm$ 9.17 & 31.03 $\pm$ 8.41 & 0.40 $\pm$ 0.37 & 2.49 $\pm$ 0.75 \\
& & 2D CNN + GRU         & 171.84 $\pm$ 131.04 & 0.65 $\pm$ 0.29 & 13.32 $\pm$ 5.69 & 31.17 $\pm$ 8.39 & 0.40 $\pm$ 0.36 & 2.56 $\pm$ 0.79 \\
& & 2D CNN + Transformer & 179.29 $\pm$ 123.41 & 0.61 $\pm$ 0.29 & 14.23 $\pm$ 6.36 & 46.07 $\pm$ 8.40 & -0.26 $\pm$ 0.57 & 4.13 $\pm$ 0.79 \\
& & 2D CNN + Mamba       & 168.27 $\pm$ 130.70 & 0.66 $\pm$ 0.30 & 13.01 $\pm$ 6.02 & 27.62 $\pm$ 7.71 & 0.52 $\pm$ 0.31 & 2.24 $\pm$ 0.76 \\

\bottomrule
\end{tabular}
}
\end{table*}

For the embedding representation, the 2D-CNN--GRU achieves the lowest RUL RMSE
of 147.08 cycles under the extended 30-cycle history. The same architecture
produces 171.84 cycles with ten consecutive cycles, supporting the wider
temporal coverage used by the RUL Expert. The selected long-term pipeline
therefore matches Section~\ref{subsec:rul_expert}: pretrained GRU embeddings are
processed by a 2D-CNN and a temporal GRU over ten cycles sampled from a 30-cycle
window.

For statistical features, the 2D-CNN--Transformer achieves 151.04 cycles RUL
RMSE and 14.41~mAh capacity RMSE using the ten most recent consecutive cycles.
With the extended history, the same backbone gives 154.43 cycles and
14.27~mAh. The capacity difference is only 0.14~mAh, while the consecutive
setting better matches the intended recent-condition role and slightly improves
RUL RMSE. It is therefore selected for the Capacity Expert. These results show
that the final experts are specialized by both representation and temporal
coverage rather than by assigning one output exclusively to each branch; both
experts remain jointly supervised for RUL and capacity during Stage~II.

\subsubsection{Individual Experts and Fusion Strategies}
The standalone RUL Expert achieves RMSEs of 147.08 cycles and 34.25~mAh,
whereas the standalone Capacity Expert achieves 151.04 cycles and 14.41~mAh
(Table~\ref{tab:main_results_verified}). Equal-weight FiLM reduces these to
143.69 cycles and 12.36~mAh, respectively. Because both experts are frozen in
Stage~III, this improvement is obtained by learning an interaction between the
fixed long-term representation $\mathbf{h}^{L}$ and short-term representation
$\mathbf{h}^{S}$ rather than by modifying either expert feature extractor.

\begin{figure}[t]
    \centering
    \includegraphics[width=\columnwidth]{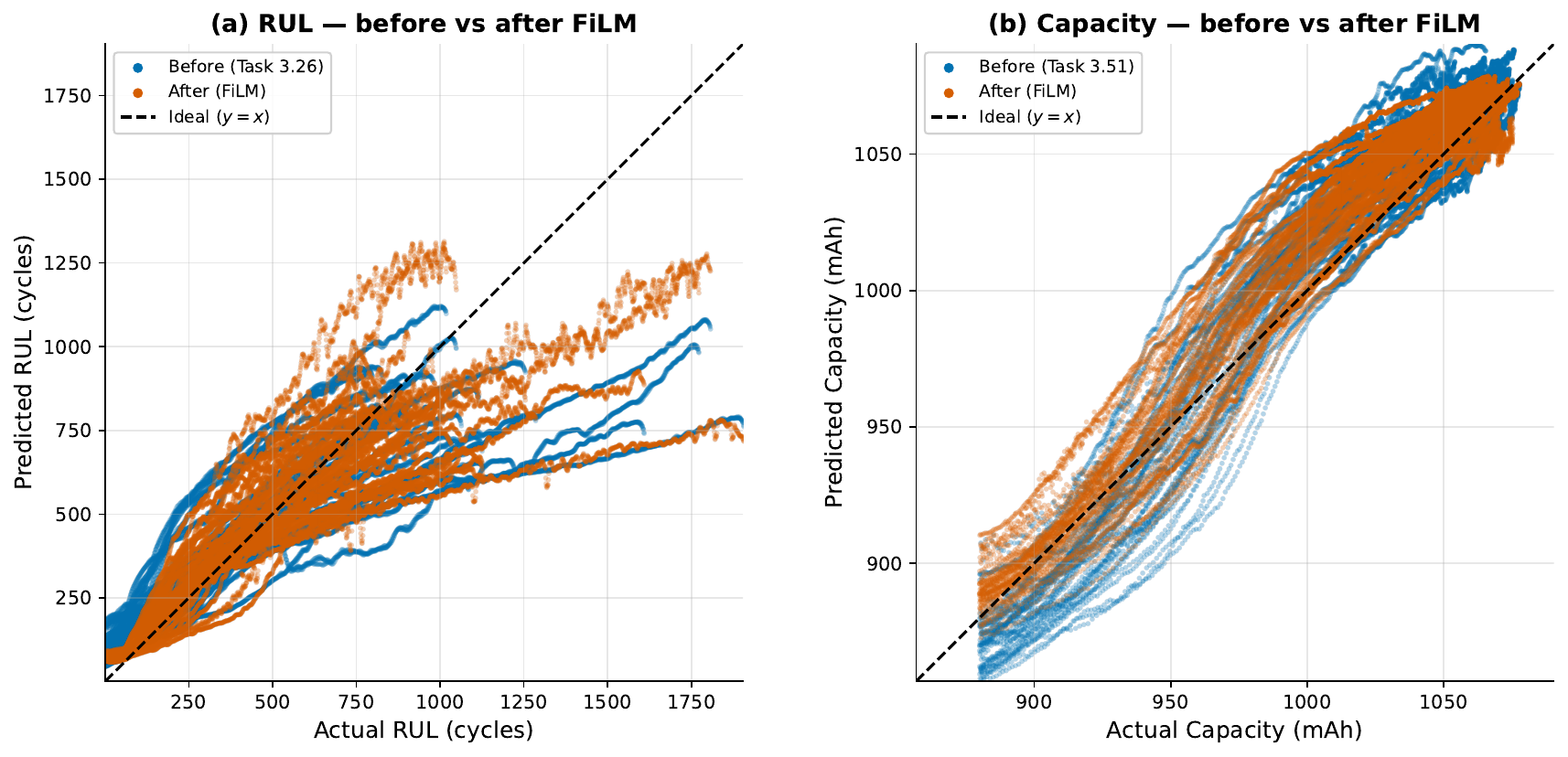}
    \caption{Predicted versus actual values before and after FiLM fusion for (a) RUL and (b) capacity on Dataset~I.}
    \label{fig:pred_vs_actual}
\end{figure}

Figure~\ref{fig:pred_vs_actual} shows the corresponding predictions before and
after FiLM. Panel~(a) uses the standalone RUL Expert as the pre-fusion reference,
whereas panel~(b) uses the standalone Capacity Expert. FiLM reduces part of the
high-RUL underestimation and moves many capacity predictions closer to the ideal
$y=x$ line, although residual bias remains and the improvement is not uniform
for every sample.

% \begin{figure}[H]
%     \centering
%     \includegraphics[width=0.8\textwidth]
%     {Images/pred_vs_actual_fusion.pdf}
%     \caption{Predicted versus actual values before and after FiLM fusion for (a) RUL and (b) capacity on Dataset~I.}
%     \label{fig:pred_vs_actual}
% \end{figure}

Table~\ref{tab:rd_fusion} compares alternative Stage-III fusion operators using
the same two frozen expert representations. Attention pooling gives the lowest
RUL RMSE (143.55 cycles), only 0.14 cycles below FiLM (143.69), while FiLM gives
the lowest RUL MAPE (10.91\%) and the best capacity metrics
(12.36~mAh, $R^2=0.90$, MAPE 0.99\%). Relative to feature concatenation and
attention pooling, FiLM reduces capacity RMSE by 18.25\% and 33.80\%,
respectively, while their RUL RMSEs remain nearly identical. FiLM is therefore
selected for the proposed framework because it provides the strongest joint-task
trade-off and directly implements the intended conditioning mechanism in
Eq.~\eqref{eq:film}: recent short-term behavior generates feature-wise scaling
and shifting terms for the long-term degradation representation.

\begin{table*}[t]
\centering
\caption{Comparison of fusion operators with frozen experts on Dataset~I. Boldface denotes the best mean value for each metric.}
\label{tab:rd_fusion}
\footnotesize
\setlength{\tabcolsep}{4pt}
\renewcommand{\arraystretch}{1.08}
\resizebox{0.8\textwidth}{!}{%
\begin{tabular}{lcccccc}
\toprule
\multirow{2}{*}{\textbf{Configuration}} & \multicolumn{3}{c}{\textbf{RUL Prediction}} & \multicolumn{3}{c}{\textbf{Capacity Estimation}}\\
\cmidrule(lr){2-4}\cmidrule(lr){5-7}
& \textbf{RMSE (cycles) $\downarrow$} &
\textbf{$R^2$ $\uparrow$} &
\textbf{MAPE (\%) $\downarrow$} &
\textbf{RMSE (mAh) $\downarrow$} &
\textbf{$R^2$ $\uparrow$} &
\textbf{MAPE (\%) $\downarrow$} \\
\midrule
Mixture of Experts (gate)~\cite{cai2024survey} &  152.02 $\pm$ 113.19 & 0.71 $\pm$ 0.26 & 11.92 $\pm$ 5.85 & 35.56 $\pm$ 9.87 & 0.22 $\pm$ 0.43 & 3.13 $\pm$ 0.92 \\
Element-wise Addition &  144.01 $\pm$ 106.53 & \textbf{0.74 $\pm$ 0.23} & 11.06 $\pm$ 5.26 & 14.94 $\pm$ 6.80 & 0.84 $\pm$ 0.13 & 1.23 $\pm$ 0.57 \\
Feature Concatenation &  143.57 $\pm$ 103.43 & 0.73 $\pm$ 0.24 & 11.32 $\pm$ 5.40 & 15.12 $\pm$ 7.50 & 0.84 $\pm$ 0.14 & 1.27 $\pm$ 0.66 \\
Attention Pooling &  \textbf{143.55 $\pm$ 103.48} & \textbf{0.74 $\pm$ 0.23} & 11.31 $\pm$ 5.30 & 18.67 $\pm$ 8.34 & 0.76 $\pm$ 0.20 & 1.59 $\pm$ 0.75 \\
Cross-Attention &  169.58 $\pm$ 129.38 & 0.64 $\pm$ 0.30 & 13.10 $\pm$ 6.15 & 30.37 $\pm$ 9.95 & 0.42 $\pm$ 0.37 & 2.65 $\pm$ 0.96 \\
Gated Multimodal Unit (GMU)~\cite{arevalo2017gated} &  144.59 $\pm$ 104.30 & 0.73 $\pm$ 0.24 & 11.37 $\pm$ 5.34 & 20.02 $\pm$ 8.47 & 0.73 $\pm$ 0.23 & 1.70 $\pm$ 0.78 \\
Bilinear &  153.80 $\pm$ 115.65 & 0.70 $\pm$ 0.25 & 12.08 $\pm$ 5.73 & 32.06 $\pm$ 9.70 & 0.36 $\pm$ 0.38 & 2.56 $\pm$ 0.91 \\
Max Pooling &  143.77 $\pm$ 104.43 & 0.73 $\pm$ 0.24 & 11.28 $\pm$ 5.42 & 18.40 $\pm$ 8.23 & 0.77 $\pm$ 0.20 & 1.57 $\pm$ 0.73 \\
Hadamard Product &  145.23 $\pm$ 107.23 & \textbf{0.74 $\pm$ 0.23} & 11.34 $\pm$ 5.15 & 44.37 $\pm$ 12.22 & -0.22 $\pm$ 0.69 & 3.74 $\pm$ 1.19 \\
FiLM (feature modulation) &  143.69 $\pm$ 107.15 & \textbf{0.74 $\pm$ 0.23} & \textbf{10.91 $\pm$ 5.17} & \textbf{12.36 $\pm$ 5.32} & \textbf{0.90 $\pm$ 0.08} & \textbf{0.99 $\pm$ 0.45} \\
\bottomrule
\end{tabular}%
}
\end{table*}

The weaker capacity performance of cross-attention, bilinear fusion, Hadamard
product, and the prediction-level mixture of experts further indicates that
simply increasing fusion complexity does not guarantee better joint prediction.
The results support FiLM as the selected representation-level interaction, but
do not by themselves identify the physical degradation factors encoded by its
modulation parameters.

\subsubsection{Capacity-Loss Weighting in Fusion Training}
Table~\ref{tab:rd_loss} evaluates the Stage-III capacity-loss weight with both experts
frozen, using the scaled RUL and capacity targets defined in
Section~\ref{subsec:training_strategy}. The reference setting
$\lambda_{\mathrm{Cap}}=1$ gives 143.69 cycles RUL RMSE and
12.36~mAh capacity RMSE. Reducing
$\lambda_{\mathrm{Cap}}$ to 0.5 lowers RUL RMSE to 141.64 cycles but increases
capacity RMSE to 15.50~mAh. Increasing $\lambda_{\mathrm{Cap}}$ to 2 does not improve capacity RMSE over
the reference setting. Thus, the response is not monotonic because the capacity-loss
weight changes the shared optimization trajectory rather than directly setting the
physical-unit error of either output.

\begin{table*}[t]
\centering
\caption{Effect of the capacity-loss weight on FiLM fusion performance on Dataset~I. Task-removal controls are also included. Boldface denotes the best mean value for each metric.}
\label{tab:rd_loss}
\footnotesize
\setlength{\tabcolsep}{4pt}
\renewcommand{\arraystretch}{1.08}
\resizebox{0.8\textwidth}{!}{%
\begin{tabular}{lcccccc}
\toprule
\multirow{2}{*}{\textbf{Configuration}} & \multicolumn{3}{c}{\textbf{RUL Prediction}} & \multicolumn{3}{c}{\textbf{Capacity Estimation}}\\
\cmidrule(lr){2-4}\cmidrule(lr){5-7}
& \textbf{RMSE (cycles) $\downarrow$} &
\textbf{$R^2$ $\uparrow$} &
\textbf{MAPE (\%) $\downarrow$} &
\textbf{RMSE (mAh) $\downarrow$} &
\textbf{$R^2$ $\uparrow$} &
\textbf{MAPE (\%) $\downarrow$} \\
\midrule
$\lambda_{\mathrm{Cap}}=1$ &  143.69 $\pm$ 107.15 & \textbf{0.74 $\pm$ 0.23} & 10.91 $\pm$ 5.17 & \textbf{12.36 $\pm$ 5.32} & \textbf{0.90 $\pm$ 0.08} & \textbf{0.99 $\pm$ 0.45} \\
RUL loss only ($\lambda_{\mathrm{Cap}}=0$) &  144.09 $\pm$ 108.54 & \textbf{0.74 $\pm$ 0.23} & \textbf{10.87 $\pm$ 5.17} & 993.27 $\pm$ 12.24 & -559.18 $\pm$ 102.52 & 93.07 $\pm$ 1.00 \\
Capacity loss only ($\mathcal{L}_{\mathrm{III}}=\mathcal{L}_{\mathrm{Cap}}$) &  778.07 $\pm$ 179.51 & -6.55 $\pm$ 0.96 & 72.99 $\pm$ 5.24 & 18.77 $\pm$ 9.02 & 0.75 $\pm$ 0.20 & 1.56 $\pm$ 0.84 \\
$\lambda_{\mathrm{Cap}}=0.5$ &  \textbf{141.64 $\pm$ 105.12} & \textbf{0.74 $\pm$ 0.23} & 10.91 $\pm$ 5.36 & 15.50 $\pm$ 6.56 & 0.84 $\pm$ 0.14 & 1.26 $\pm$ 0.56 \\
$\lambda_{\mathrm{Cap}}=2$ &  144.63 $\pm$ 107.08 & \textbf{0.74 $\pm$ 0.23} & 11.22 $\pm$ 5.17 & 12.85 $\pm$ 5.96 & 0.89 $\pm$ 0.10 & 1.06 $\pm$ 0.52 \\
\bottomrule
\end{tabular}%
}
\end{table*}

The $\lambda_{\mathrm{Cap}}=1$ configuration is retained as the reference joint-task model
because it provides the lowest capacity error among the jointly supervised
weight settings while remaining close to the minimum RUL error. The
$\lambda_{\mathrm{Cap}}=0.5$ result is reported separately when RUL accuracy is
prioritized. Removing either task loss severely degrades the unsupervised
output, confirming that both outputs require direct supervision in the shared
Stage-III head; this observation alone should not be interpreted as proof of
positive transfer between the tasks.

\subsection{Computational Considerations}

\begin{table}[H]
\centering
\caption{Computational efficiency of the proposed method and simple baselines.}
\label{tab:efficiency}
\begin{tabular}{lccc}
\toprule
\textbf{Method} & \textbf{Params (M)} & \textbf{Inference (ms)} & \textbf{Memory (MB)} \\
\midrule
GRU         & 0.159 & \textbf{8.98} & 1100.4 \\
LSTM        & 0.209 & 18.85 & 1416.2 \\
Transformer & 0.406 & 35.31 & \textbf{518.5} \\
\midrule
Proposed method & 1.103 & 29.54 & 1300.9 \\
\bottomrule
\end{tabular}
\end{table}

Table XI reports inference for a single model instance; reproducing the five-model ensemble used for the reported predictive results requires five model evaluations. The proposed framework contains more parameters than the simple baselines
because it retains two complementary expert branches and the FiLM-based
fusion module. However, its three-stage training strategy limits the number
of parameters optimized simultaneously. In Stage~I, the autoencoder and
auxiliary predictor are pretrained; in Stage~II, the transferred encoder is
frozen while the remaining RUL Expert and Capacity Expert components are
trained independently; and in Stage~III, both experts are frozen so that only
the FiLM module and shared regression head are optimized. Thus, the larger
total model size does not imply that all parameters are updated during the
final fusion stage.

At inference, the Stage-I decoder and auxiliary prediction backbone are
discarded, while the RUL Expert, Capacity Expert, FiLM module, and shared
regression head are retained. As shown in Table~\ref{tab:efficiency}, the
proposed method requires 29.54~ms per batch, which is slower than the GRU
and LSTM baselines but faster than the Transformer baseline. Its peak memory
usage of 1300.9~MB is higher than that of the GRU and Transformer baselines
but lower than that of the LSTM.

\subsection{Limitations and Future Work}
The two datasets contain LFP/graphite cells tested under controlled
ambient conditions. The current results therefore demonstrate
performance on separately trained datasets rather than zero-shot
cross-dataset transfer or validation under field conditions. Future
work will evaluate the proposed framework across additional battery
chemistries, temperatures, and operating policies, and will investigate
domain adaptation and transfer-learning strategies for deployment under
previously unseen operating conditions.

Although segment-local charge integration avoids the need for measured
historical full-cycle capacity, the framework still requires sufficient
partial-charging observations and a predefined cycle history. Future
work will therefore investigate more flexible handling of incomplete or
interrupted charging segments, sensor noise and missing measurements,
and charging sessions that begin above the selected voltage threshold.
In addition, the current Capacity Expert requires a nominal 40-min
charging segment, which may not be available in every operating cycle.
Reducing this observation requirement while preserving joint RUL and
capacity accuracy will be an important direction for practical BMS
deployment.

\section{Conclusion}
\label{sec:conclusion}
This paper presents a three-stage framework for joint RUL prediction and capacity estimation from partial-charging observations. A GRU-autoencoder-based long-term view and a statistical short-term view are processed by specialized temporal backbones and combined through FiLM, with the short-term representation conditioning the long-term representation. The reference configuration achieves mean RUL RMSEs of 143.69 and 161.10 cycles and capacity RMSEs of 12.36 and 7.28~mAh on Datasets~I and II, respectively. On Dataset~I, equal-weight fusion improves both mean errors relative to the standalone experts. A capacity-loss weight of 0.5 further reduces RUL RMSE to 141.64 cycles but increases capacity RMSE to 15.50~mAh. The results support complementary feature fusion while showing that the proposed method does not minimize capacity error across all baselines. Future work will examine broader operating conditions, incomplete charging observations, uncertainty calibration, and measured deployment costs.

\section*{Data and Code Availability}
The datasets are described in \cite{severson2019data,ma2022real}.
Code inquiries may be directed to AIWARE Limited Company at https://aiware.website.

\section*{Preprint}
A preprint version of this manuscript is available at https://arxiv.org/abs/2609.21932.

\bibliographystyle{plain}
\bibliography{cas-refs}

\end{document}